\documentclass{article}
\usepackage{arxiv}

\usepackage{hyperref}
\usepackage{bm}% allocate bold fonts
\usepackage{cite}
\usepackage{graphics,fancyhdr,graphicx,subfigure}
\usepackage{subfigure}
\usepackage{amsthm,amsmath,amsfonts,latexsym,amssymb}
\usepackage{lineno}
\usepackage{diagbox}
\usepackage{algorithm}
\PassOptionsToPackage{noend}{algpseudocode}
\usepackage{algpseudocode}
\usepackage{multirow,booktabs}
\usepackage{listings}
\usepackage[table]{xcolor}
\usepackage{lscape}
\usepackage{comment}
\usepackage{tabularx}
\usepackage{graphicx}%to add images
\usepackage{tabularx}
\usepackage{arydshln}
\usepackage{nomencl}%for adding symbols page
\usepackage{array}%to add table
\usepackage{caption}%add caption for the table and list it in list of tables
\usepackage{breqn}
\usepackage{lineno}
\usepackage{mathtools}
\usepackage{subfigure}
\usepackage{caption}
\usepackage{lmodern}
\usepackage{calligra}
\usepackage{xspace}
\usepackage[utf8]{inputenc}%skiplength
\usepackage{enumerate}%list
\usepackage[english]{babel}
\def\equationautorefname~#1\null{%
  Eq.~(#1)\null
  }
\def\subfigureautorefname~#1\null{%
  Fig.~#1\null
}

\definecolor{listinggray}{gray}{0.9}
\definecolor{lbcolor}{rgb}{0.9,0.9,0.9}
\definecolor{Darkgreen}{RGB}{0,100,0}

\title{Enhanced multi-fidelity modelling for digital twin and uncertainty quantification \thanks{https://www.csccm.in/}}

\author{ \hspace{1mm}Desai~A S.{$^\dagger$}\\
	Department of Applied Mechanics\\
	Indian Institute of Technology (IIT) Delhi\\
	Hauz Khas - 110 016, New Delhi, India \\
	\texttt{Aarya.Sheetal.Desai.ama20@am.iitd.ac.in} \\
	\And
        \hspace{1mm}Navaneeth~N.{$^\dagger$}\\
	Department of Applied Mechanics\\
	Indian Institute of Technology (IIT) Delhi\\
	Hauz Khas - 110 016, New Delhi, India \\
	\texttt{navaneeth.n@am.iitd.ac.in} \\
	\And
        \hspace{1mm}Sondipon~Adhikari\\
	James Watt School of Engineering\\
	The University of Glasgow\\
	Glasgow G12 8QQ, UK\\
	\texttt{Sondipon.Adhikari@glasgow.ac.uk} \\
	\And
	\hspace{1mm}Souvik~Chakraborty \\
	Department of Applied Mechanics,\\
        Yardi School of Artificial Intelligence\\
	Indian Institute of Technology (IIT) Delhi\\
	Hauz Khas - 110 016, New Delhi, India \\
	\texttt{souvik@am.iitd.ac.in} \\
}

\renewcommand{\shorttitle}{Enhanced multi-fidelity modelling for digital twin and uncertainty quantification}

\hypersetup{
pdftitle={Enhanced multi-fidelity modelling for digital twin and uncertainty quantification},
pdfsubject={q-bio.NC, q-bio.QM},
pdfauthor={Desai A S, Navaneeth N., Sondipon Adhikari, Souvik Chakraborty},
pdfkeywords={Multi-fidelity, Deep-H-PCFE, uncertainty quantification, Surrogate models,
digital twin},
}
\begin{document}
\maketitle

\begin{abstract}
The increasing significance of digital twin technology across engineering and industrial domains, such as aerospace, infrastructure, and automotive, is undeniable. However, the lack of detailed application-specific information poses challenges to its seamless implementation in practical systems. Data-driven models play a crucial role in digital twins, enabling real-time updates and predictions by leveraging data and computational models. Nonetheless, the fidelity of available data and the scarcity of accurate sensor data often hinder the efficient learning of surrogate models, which serve as the connection between physical systems and digital twin models. To address this challenge, we propose a novel framework that begins by developing a robust multi-fidelity surrogate model, subsequently applied for tracking digital twin systems.
Our framework integrates polynomial correlated function expansion (PCFE) with the Gaussian process (GP) to create an effective surrogate model called H-PCFE. Going a step further, we introduce deep-HPCFE, a cascading arrangement of models with different fidelities, utilizing nonlinear auto-regression schemes. These auto-regressive schemes effectively address the issue of erroneous predictions from low-fidelity models by incorporating space-dependent cross-correlations among the models. To validate the efficacy of the multi-fidelity framework, we first assess its performance in uncertainty quantification using benchmark numerical examples. Subsequently, we demonstrate its applicability in the context of digital twin systems. 
\end{abstract}


\def\thefootnote{$\dagger$}\footnotetext{These authors contributed equally to this work}\def\thefootnote{\arabic{footnote}}

\section{Introduction}
A digital twin (DT) is described as a cloud-based digital counterpart of a physical system that communicates with the original system via the Internet of Things (IoT). The digital twin technique differs from the other computational models owing to the fact that the digital twin model is updated continuously in order to achieve synchronization with the actual system as it evolves. The synchronization is achieved through a data acquisition system, machine learning modules, and the Internet of Things associated with the digital twin.
The notion of the digital twin is so broad and has a wide range of industrial applications, including smart manufacturing, mechanical and aerospace industries, smart cities, and urban planning. Rather than adopting a unified approach, the implementation of digital twins might differ in their approach to the specifics of the  application. All digital twin models start as  nominal models; however, with the advancement in time, it is essential to track the evolution of the physical systems and to make key decisions at a point in time in the future. For that, the digital twin is trained to synchronize with the actual physical systems.

Considering the history of digital twins, the concept of a digital twin has been discussed in numerous works over the past two decades. A general mathematical foundation for digital twins is introduced in \cite{worden2020digital}. Ensuing noteworthy studies that have contributed to the development of the digital twin include the investigation of the digital twin for prognostics and health monitoring \cite{worden2020digital,booyse2020deep,tao2019digital}, manufacturing \cite{haag2018digital, lu2020digital,zhang2019reconfigurable}, and automotive and aerospace engineering \cite{li2017dynamic,kapteyn2020toward,tuegel2011reengineering}. While the definition of a digital twin is widely accepted, there is almost no formal method for developing one for a specific system. A physics-driven digital twin is one of the plausible approaches \cite{ganguli2020digital}; nonetheless, which is liable to some significant drawbacks. One of the fundamental challenges is that it can provide only state estimates at discrete time steps. Secondly, such digital twin fails to yield accurate results when the measured data contains complex noises. This motivates researchers  to consider surrogate modeling-based approaches \cite{chakraborty2021role,barkanyi2021modelling,adeyemo2022surrogate,yu2022hybrid,chakraborty2021machine,ganguli2023digital} as an efficient alternative approach to modelling digital twins.

A surrogate model can be defined as a substitute for an actual high-fidelity model, which is often a time-consuming computational model or an expensive experimental model. While there are several surrogate-based approaches available in the literature, the popular representative ones include polynomial regression surface (PRS) \cite{shanok,han2016kriging}, artificial neural networks (ANN) \cite{1000125,yan2020efficient, khorrami2023artificial}, radial basis function (RBF) \cite{gutmann,liu2022adaptive}, Gaussian process (GP) \cite{bilionis2013multi,chakraborty2019graph,hashemi2023gaussian}, and support vector regression (SVR) \cite{andres2012efficient,roy2023support}. More extensive studies and comparisons of these approaches are presented in \cite{wang2006review}.  In general, the success of the surrogate models is recognized by the requirement of the volume of data  to train the model efficiently. While on the one hand, smaller data leads to inefficient training, on the other hand,  a larger data set results in  a greater computational cost. Consequently, the trade-off between accuracy and cost is always challenging for the data-driven surrogate models trained with high-fidelity data samples. One obvious approach is to rely on both high-fidelity data/multi-fidelity data, along with low-fidelity data, where the low-fidelity data is obtained through relatively simpler procedures. However, in that case, training with data samples having multiple fidelity data becomes the major hurdle for most of the surrogate-based approaches.

Researchers have considered the multi-fidelity modelling approach as a viable approach that can circumvent the aforementioned disadvantages associated with surrogate models. While there are several multi-fidelity modelling approaches available in the literature \cite{zhang2023multi, zhang2022multi, thakur2022multi, shang2023efficient}, many of these multi-fidelity approaches are devised as a combination of the linear auto-regressive information fusion scheme \cite{kenn10.2307/2673557,jin2021combining} and the Gaussian process (GP) regression \cite{williams2006gaussian,romor2021multi}. These seminal multi-fidelity modelling approaches employ a linear mapping between the fidelities by utilizing the cross-correlation of data having different fidelity levels within the surrogate model. However, these methods fail to achieve the desired performance when the fidelity of the data exhibits a more complex nonlinear relationship. To that end, we propose a multi-fidelity surrogate modelling approach that can efficiently model the digital twin reliant on multi-fidelity measurements.

The proposed framework is named deep Hybrid Polynomial Correlation Function Expansion (Deep H-PCFE) as the model results from the composition of individual H-PCFE corresponding to the different fidelities. As a regression model, H-PCFE  marriages the benefits of the Polynomial Correlated Function Expansion (PCFE) and the Gaussian Process (GP) \cite{chatterjee2016bi,navaneeth2022surrogate}. Since PCFE deals with dependent and independent random variables without the need for any ad-hoc transformations, PCFE becomes a computationally efficient regression model for learning the global behaviour of the function, and GP effectively captures the local variations. It is, therefore, reasonable to infer that composing the H-PCFE corresponding to the individual fidelity utilizing nonlinear cross-correlations prospectively results in a novel paradigm for multi-fidelity modelling.

While the proposed framework can effectively fusion the data of different fidelities, we first elucidate the performance of the deep-H-PCFE as a multi-fidelity surrogate model with numerical examples for the applications of uncertainty quantification and then employ the framework for digital twins. The scope of the work can have a significant impact, especially on the application of digital twins. The sensor data collected from the physical system is indispensable for updating the digital twin model. Direct field measurement utilizing accurate sensors obtains high-fidelity data, however, they are expensive. On the other hand, automated data acquisition systems and less precise sensors yield low-fidelity data. Considering this practical aspect, a robust multi-fidelity surrogate model is essential for the real-time modelling of digital twins. 

%multifididleyt data for digital twin
%It helps to reduce the cost
%it requires minimal high-fidelity data  
%field inspection is difficult

The remainder of the paper is as follows. The proposed approach is described in Section 2. Section 3 discusses the details of Deep H-PCFE as multi-fidelity surrogate modelling. In Section 4, the effectiveness of the proposed approach is demonstrated through numerical examples,  which include the application of uncertainty quantification. Subsequently, the application of the deep H-PCFE for the digital twin is illustrated in Section 5. Finally, the concluding remarks are given in Section 6.

\section{Proposed approach}\label{proposedapparoach}
The section provides a detailed description of the proposed multi-fidelity approach. The overall framework is achieved through recursive surrogate modelling with Hybrid-Polynomial Correlated Function Expansion (H-PCFE) being the basic building block. Before elaborating on the proposed approach, we first review the H-PCFE. Note that the discussion pertaining to the development of digital twins using the proposed approach is not discussed in this section.

\subsection{Hybrid-Polynomial Correlated Function Expansion (H-PCFE)}\label{HPCFE}
H-PCFE was developed by integrating the benefits of two existing methodologies, namely PCFE and the Gaussian process (GP). The PCFE capture the global behaviour of the model using a set of component functions, whereas GP interpolates local variation within  the data points, resulting in a two-stage estimate \cite{chatterjee2016bi}. Regarding PCFE, which is a surrogate modelling method that employs correlated functions as basis functions\cite{PCFE, HDMR}. Since the method effectively handles both 
dependent and independent random variables without the requirement for any ad-hoc transformations, it necessitates substantially less computing effort \cite{PCFE}. Now to describe mathematically, we suppose, \textbf{i} = $({i_1},{i_2},....,{i_N}) \in N_0^N$ with $|\textbf{i}|$ = $({i_1}+{i_2},....+{i_N})$ where $N \ge 0$ is an integer and $\bm {x} = ({x_1},{x_2},....,{x_N})$ is an N-dimensional input vector. Thus the PCFE of $f(\bm{x})$ can be represented as,
\begin{equation}
  f(\bm{x}) = \sum\limits_{|\mathbf i| = 0}^{N} {f_{\mathbf{i}}({x}_{\mathbf{i}})}\label{pcfe}
\end{equation}
In the PCFE, as indicated in Eq. \eqref{pcfe}, the first-order component functions or the univariate terms, indicate the independent influence of input variables. On the other hand, second-order component functions are bivariate, and that reflects the cooperate impact of variables. The noteworthy fact here is that the order in PCFE denotes the interaction between variables rather than the highest order of polynomials. The first-order PCFE estimate, for instance, does not reflect the linear variation and comprises terms with a higher degree of nonlinearity. To provide further elaboration, we consider $\psi$ as suitable basis for $\bm{x}$ and $\alpha$ being unknown coefficients. Thus, PCFE of $f(\bm{x})$ with mean response ${f_0}$ can be rewritten as follows:
\begin{equation}\label{pcfe_eq}
    f(\textbf{x}) = {f_0} + \sum\limits_{k = 1}^M {\{ \sum\limits_{{i_1} = 1}^{N - k + 1} {....\sum\limits_{{i_k} = {i_{k - 1}}}^N {\sum\limits_{r = 1}^k {[\sum\limits_{{m_1} = 1}^s {\sum\limits_{{m_2} = 1}^s {....\sum\limits_{{m_r} = 1}^s {\alpha _{{m_1}{m_2}....{m_r}}^{({i_1}{i_2}...{i_k}){i_r}}\psi _{{m_1}}^{{i_1}}....\psi _{{m_r}}^{{i_r}}} } } ]} } } \} } 
\end{equation}
PCFE has certain key  features that make it an ideal candidate for regression problems. One of the primary salient features is the convergence of  PCFE in the mean square sense. All the unknown parameters associated with the bases are obtained by minimizing the $L^2$ error norm. Secondly, since PCFE is a converging series made up of $2^N$ component functions, if the component functions are convergent, PCFE yields a convergent solution. Lastly, PCFE is optimum in the Fourier sense because,i.e., apart from reducing the $L^2$ error norm, it also assures that the component functions are orthogonal. Thus, in essence, PCFE provides a good global approximation. However, it falls short of accurately capturing local variations, and hence the PCFE may not perform well with functions that are locally oscillating, which is regarded as one of the fundamental limitations of the method. To alleviate this limitation, we employ Hybrid-PCFE (H-PCFE), which incorporates GP with PCFE. We delve into the details of H-PCFE  herein onwards.
For the input variable, $\bm{x}=\left(x_{1}, x_{2}, \ldots, x_{N}\right)$: $\bm{x} \in D \subset \mathbb{R}^{N}$, the output, $\mathbf{M}^{(\mathbf{H}-\mathbf{P C F E})}$ can be represented as
\begin{equation}\label{hpcfe_main}
    f_{\texttt{H-PCFE}}\approx {\tilde{f}_{\texttt{H-PCFE}}}=f_{0}+\sum_{k=1}^{M}\Bigg\{\sum_{i_{1}=1}^{N-k+1}....\sum_{i_{k}=i_{k}-1}^{N}\sum_{r=1}^{k}\Bigg(\sum_{m_{1}=1}^{b}....\sum_{m_{r}=1}^{b}{\alpha}_{m_{1}....m_{r}}^{({i_{1}}{i_{2}}...{i_{k}}){i_{r}}}{{\psi}_{m_{1}}^{i_{1}}....{\psi}_{m_{r}}^{i_{r}}}\Bigg)\Bigg\} + f_{\mathrm{GP}}
\end{equation}
In Eq. \eqref{pcfe_eq}, $\bm{\psi}$ indicate basis functions, and $\bm \alpha$ are unknown coefficients associated with the bases, where $f_0$ represents the mean response. Though any basis function can be used for constructing extended bases for PCFE, orthogonal basis functions are chosen for faster convergence. The last term of Eq. \eqref{hpcfe_main}, i.e., $f_{\mathrm{GP}}$, represents GP with zero mean. Mathematically, GP with a zero mean is expressed as:
\begin{equation}
    f_{\mathrm{GP}}=\sigma^{2} Z(\mathbf{0}, k(\cdot, \cdot ; \bm{\theta}))
\end{equation}
where $k(\cdot, \cdot ;\bm{\theta})$ is covariance kernel function with $\bm{\theta}$ as length scale parameter and $\sigma^{2}$ as process variance. In H-PCFE, the unknown parameters are $\bm{\theta},\alpha,$ and $\sigma^{2}$, which are computed by maximizing the log-likelihood,
\begin{equation}
f_{\mathrm{ML}}=\frac{1}{m} \log |\mathbf{R}(\theta)|+\log \left(\boldsymbol{d}^{T} \mathbf{R}(\theta)^{-1} \bm {d}\right)
\label{theta}
\end{equation}
This further yields the variance ${\sigma}^{2}$ as,
\begin{equation}
\sigma^{2}=\frac{1}{N_{s}}(\bm{d}-\bm{\Psi} \bm{\alpha})^{T} \mathbf{R}^{-1}(\bm{d}-\bm{\Psi} \boldsymbol{\alpha}),
\end{equation}
and weighted normal equations as
\begin{equation}
\left(\bm{\Psi}^{T} \mathbf{R}^{-1} \bm{\Psi}\right) \bm{\alpha}=\bm{\Psi}^{T} \mathbf{R}^{-1} \bm{d}.
\label{AB}
\end{equation}
Here basis function matrix, $\bm{\Psi}$, is formed from input variables. While $\mathbf{R}$ is the covariance matrix formed by training inputs and covariance kernel $k(\cdot, \cdot; \bm \theta)$, $\bm d$ is a difference between response at training inputs and mean response, i.e., $\bm d=\bm y - f_{0}$. By substituting the $\mathbf{C}=\left(\bm{\Psi}^{T} \mathbf{R}^{-1} \bm{\Psi}\right)$ and $\mathbf{D} = \bm{\Psi}^{T} \mathbf{R}^{-1} \bm{d}$, Eq. \eqref{AB} can be rewritten as, $\mathbf{C} \alpha=\mathbf{D}$. However, since matrices $\mathbf{C}$ and $\mathbf{D}$ contain extended bases, there are redundant rows; removal of the redundant rows results in $\mathbf{C'} \alpha=\mathbf{D'}$. This leads to an under-determined set of equations. The least-square solution of the system of equations is given by:
\begin{equation}
\boldsymbol{\alpha}_{0}=\mathbf{C}^{\prime \dagger} \boldsymbol{D}^{\prime}
\label{redundants}
\end{equation}
The equation above minimizes the $L^{2}$-norm such that $\mathbf{C}^{\prime \dagger}$ yields the pseudo inverse of $\mathbf C'$. In the context of Eq. \eqref{redundants}, the optimal solution is the one that minimizes the least-squared error and guarantees the hierarchical orthogonality of the component functions \cite{hp}. To that end, the homotopy algorithm is conveniently employed here to compute the unknown coefficients \cite{li2010d}. To enforce the hierarchical orthogonality condition, in the homotopy algorithm, an additional objective function is imposed through the weight matrix $\mathbf{W}_{H A}$. Therefore, the final solution achieved as follows:
\begin{equation}
\label{final}
 \boldsymbol{\alpha}_{\mathbf{H A}}=\left[\mathbf{V}_{q-r}\left(\mathbf{U}_{q-r}^{T} \mathbf{V}_{q-r}\right)^{-1} \mathbf{U}_{q-r}^{T}\right] \boldsymbol{\alpha_{0}} 
\end{equation}
where $\mathbf{U}$ and $\mathbf{V}$ are obtained by singular decomposition of $\mathbf{P} \mathbf{W}_{H A}$. 
\begin{equation}
\begin{aligned}
&\mathbf{P} \mathbf{W}_{H A}=\mathbf{U}\left[\begin{array}{cc}
\mathbf{D}_{\mathbf{r}} & 0 \\
0 & 0
\end{array}\right] \mathbf{V}^{T} \\
&\mathbf{P}=\left[\mathbb{I}-\left(\mathbf{A}^{\prime}\right)^{-1} \mathbf{A}^{\prime}\right]
\end{aligned}
\label{PIA}
\end{equation}
Here we note that in the equation, Eq. \eqref{final}, $\mathbf{V}_{q-r}$ $\mathbf{U}_{q-r}$ are last $(q-r)$ rows of $\mathbf{V}$ and $\mathbf{U}$.
For a detailed explanation of the weight matrix in the above formulations, interested readers may refer to \cite{li2012d} and for details on HA, follow \cite{li2010d}. Once $\boldsymbol{\alpha}_{HA}$ is obtained, at any unknown point, predictive mean and variance can be calculated as,
\begin{equation}
\begin{aligned}
&\mu\left(\boldsymbol{x}^{*}\right)=f_{0}+\boldsymbol{\Phi}\left(\boldsymbol{x}^{*}\right) \boldsymbol{\alpha}_{H A}+\boldsymbol{r}\left(\boldsymbol{x}^{*}\right) \mathbf{R}^{-1}\left(\boldsymbol{d}-\boldsymbol{\Psi}_{\boldsymbol{H} A}\right) \\
&s^{2}\left(\boldsymbol{x}^{*}\right)=\sigma^{2}\left\{1-\boldsymbol{r}\left(\boldsymbol{x}^{*}\right) \mathbf{R}^{-1} \boldsymbol{r}\left(\boldsymbol{x}^{*}\right)^{T}+\frac{1-\boldsymbol{\Psi}^{T} \mathbf{R}^{-1} \boldsymbol{r}\left(\boldsymbol{x}^{*}\right)^{T}}{\boldsymbol{\Psi}^{T} \mathbf{R}^{-1} \boldsymbol{\Psi}}\right\}
\end{aligned}
\end{equation}
In the above expression, $\bm \Phi\left(\bm x^{*}\right)$ is basis function vector evaluated at $\bm x^{*}$ and $\bm r(\bm x^{*})$ denotes the correlation
between training inputs and $\bm x^{*}$. A step-by-step implementation of the H-PCFE is illustrated in the Alg. \ref{alg:hpcfe}

\begin{algorithm}[ht!]
\caption{H-PCFE}
\label{alg:hpcfe}
{\textbf{Requirements:} Training data set, input order of H-PCFE,
corresponding parameters and variable bounds.}\\
\textbf{Output:} The set of unknown coefficients of the basis functions $\alpha$ in \autoref{hpcfe_main}.
\begin{algorithmic}[1]
\State Calculate $f_{0}$ from:\newline
        $f_{0}\leftarrow \frac{1}{n}\sum_{n}f({x}_{i})$
\State Using the training output data($y$), formulate $\bm{d}$ as\newline
        $\bm d\leftarrow {\bm y}-f_{0}$\newline
 $\bm{d}\leftarrow [d_{1}\: d_{2}\:... \:d_{n}]^{T}$ 
\State Calculate basis function matrix ${\Psi}$
\State Select an appropriate functional form for covariance matrix $\mathbf {R}$.
\State to obtain the length scale parameter $\theta$, We maximize the likelihood estimate{\Comment{Eq. \eqref{theta}}}
\State $\mathbf{C}\leftarrow (\mathbf{\Psi}^{T}\mathbf{R}^{-1}\mathbf{\Psi})$, $\mathbf{D}\leftarrow \mathbf{\Psi}^{T}\textbf{R}^{-1}\bm{d}$
\State Obtain $\mathbf{C}'$ and $\mathbf{D'}$ by removing redundant in matrices $\mathbf C$ and $\mathbf D$ {\Comment{Eq. \eqref{redundants}}}
\State $\bm{\alpha}_{0} \leftarrow (\mathbf{C'})^{\dagger}\mathbf{D'}${\Comment{Eq. \eqref{redundants}}}
\State $\mathbf{P}\leftarrow {\mathbf{I}}-(\mathbf{C'})^{-1}\mathbf{C'}${\Comment{Eq. \eqref{PIA}}}
\State Using homotopy algorithm weight matrix $\mathbf{W}_{HA}$ is formed.
\State From the singular value decomposition of $\mathbf{P}\mathbf{W}_{HA}$ , $\mathbf{U}$ and $\mathbf{V}$ are obtained.
\State $\bm{\alpha}_{HA} \leftarrow[\mathbf{V}_{q-r}(\mathbf{U}_{q-r}^{T}\mathbf{V}_{q-r})^{-1}\mathbf{U}_{q-r}^{T}]\bm{\alpha}_{0}${\Comment{Eq. \eqref{final}}}
\end{algorithmic}
\end{algorithm}

\subsection{Deep H-PCFE as multi-fidelity surrogate modelling}
So far, we have discussed H-PCFE as a surrogate model which learns the mapping between the paired input and output. This section discusses a systematic approach for constructing a multi-fidelity model utilizing H-PCFE. As the name implies, the multi-fidelity concept involves information of two or more levels of fidelity. Multi-fidelity models strive to provide the most accurate model predictions by the effective fusion of data of different fidelity; such that model training requires fewer high-fidelity/high-cost data points. It is generally observed that the performance of multi-fidelity models is enhanced when the low and high-fidelity data are strongly correlated \cite{article_multi}. While there exist several information fusion methods, here we adopt a recursive modeling approach  due to two significant advantages:
Firstly, the successive model learns the correlation of lower fidelity data with higher fidelity data through the prediction of the previous model. As a result, these models require lesser high-fidelity data points, which reduces computational expenses. 
Secondly, it also eliminates the erroneous results solely obtained by low-fidelity data. 

The notion of recursive prediction was inspired by nonlinear autoregression methods \cite{frf}, in which models are commonly used to predict successive time responses. In the following section, we provide a brief overview of nonlinear autoregression schemes to motivate multi-fidelity modelling using the information fusion technique.

\subsubsection{Non linear auto-regression}
The auto-regression enables the model prediction at the current variable state based on the output corresponding to the previous variable state. As a pedagogical example, consider the case of any discrete dynamical system having time series response data with $y_t$ as output at time t, and $y_{t-1}$ at $t-1$ time and so on, then a general nonlinear auto-regression  scheme can be represented as:
\begin{equation}\label{yeq}
    {y_t} = F({y_{t - 1}}, {y_{t - 2}}, {y_{t - 3}}...) + {\varepsilon _t},
\end{equation} 
where $y_t$ is output prediction at the time $t$ and $F$ is a nonlinear mapping function. $\varepsilon _t$ in the \autoref{yeq} represents the error associated with the surrogate approximation of the nonlinear function. For further calculation, $\varepsilon_t$ can be treated as a random variable with a specified probability distribution. As $F$ learns the mapping between $y_{t-1}$ and $y_t$, we can yield the prediction of $y_t$ using $y_{t-1}$ and $y_{t-2}$...$y_1$. Following the analogy, the approximation of the multi-fidelity model of the function $F$ can be achieved such that the output at $i{th}$ level i.e $y_i$ is estimated from $y_{i-1}$ and $y_{i-2}$...$y_1$, which corresponds to output at $(i-1)^{th}$, $(i-2)^{th}$...$1^{st}$ fidelity levels.

\subsubsection{Deep H-PCFE}\label{dphpc}
Consider the data having n number of fidelity levels, $\{\bm x^{1},y^{1}\}$, $\{\bm x^{2},y^{2}\}$,$......\{\bm x^{M},y^{M}\}$, such that index varies from $1$ to $M$ represents the  level of fidelity data  which varies from lowest fidelity to highest fidelity. With multiple level of mapping functions $f_i$s  with $i = 1,2,3,...,M$, the recursive surrogate modelling is formulated as:
\begin{equation}
    f_1(x)=g_1(\bm x^{1})
\end{equation}
\begin{equation}
    f_2(x)=g_2(\bm x^{2},f^{*}_{1}(\bm x^{2}))
\end{equation}
\begin{equation}
    f_3(x)=g_3(\bm x^{3},f^{*}_{2}(\bm x^{3}),f^{*}_{1}(\bm x^{3}))
\end{equation}
continuing further in this manner till $i=M$.
\begin{equation}
    f_M(x)=g_M(\bm x,f^{*}_{M-1}(\bm x^{M}),f^{*}_{M-2}(\bm x^{M}),f^{*}_{M-3}(\bm x^{M}).....f^{*}_1(\bm x^{M}))
\end{equation}
Here, there are M equations having each $f_i$ depending on the previous $i-1$ number of outputs predicted by $f$. The outputs of $f_i$ are obtained at each level fidelity through $g_i$s, where $g_{i}$ represents the trained H-PCFE model described in the section \autoref{HPCFE} with $\bm x^{i}$ being the input and $y^{i}$ being the output. In the first level, we train the $g_1$, the H-PCFE model with the lowest fidelity data. Subsequently, we train $g_2$ with input as $x^{2}$ and prediction of previous model $f_1(x^{2})$ with H-PCFE method. Similarly, for training $g_i$ we use $x^{i}$ as the input along with the output prediction of all the previous $i-1$ models at $x_i$. Since we utilize the predictions of all lower-level models as input to the highest fidelity  model, it is referred to as deep H-PCFE.

In order to comprehend the prediction with Deep H-PCFE having M-level fidelity, let us consider the test data $\bm x^*$. We first provide the data to the trained low fidelity mapping function $g_1(\bm x^*)$=$f_1(\bm x^*)$. Once the prediction $f_1(\bm x^*)$ are obtained, we concatenate it with the $\bm x^*$, i.e. [$\bm x^*$,$f_1(\bm x^*)$] passes to $g_2$. Once the prediction of $g_2$,
$f_2(\bm x^*)$ is obtained  as $f_2(\bm x^*)=g_2(\bm x^*,f_1(\bm x^*))$, $f_2(\bm x^*)$ along with the $\bm x^*$ and $f_1(\bm x^*)$ are provided to $g_3$  as input. The procedure repeats for every subsequent prediction till ${M}^{th}$ level fidelity. Thus finally the prediction of $f_M(\bm x^*)$ can be represented as $f_M(\bm x^*) = g_M(\bm x^*,f_{M-1}(\bm x^*),f_{M-2}(\bm x^*),.....,f_1(\bm x^*))$. The training algorithm is described in Alg.\ref{alg:hpcfe_train}, while the testing algorithm is demonstrated in Alg.\ref{alg:hpcfe_pred}. Here we note that, in the proposed multi-fidelity approach, we conveniently assume the mean function corresponding to the H-PCFE corresponding to the certain fidelity level as zero. This case with the deep H-PCFE is referred as modified Deep-H-PCFE. We have used the modified Deep H-PCFE approach in some of the numerical examples below, such as \autoref{buckling_pl}, where actual H-PCFE is employed at the lowest fidelity level, i.e. only $g_1$, and for the rest of the models ($g_i$ where $i \in$ [2 to M]), H-PCFE with mean function zero is used. A step-by-step implementation of deep H-PCFE is illustrated in the Alg.\ref{alg:hpcfe_train} and Alg.\ref{alg:hpcfe_pred}.

\begin{algorithm}[ht!]
\caption{deep H-PCFE training}\label{alg:hpcfe_train}
\textbf{Requirements:} Training data set consisting M number of output vectors and input vectors
\begin{algorithmic}[1]
\State{Declare $prev$ as a null matrix}
\For{$i \gets 1 $ to $ M \do$}
\State{$\bm x^i$ are $i^{th}$ fidelity inputs, $y^i$ is corresponding output at $i^{th}$ fidelity level.}
\State{Train $g_i$ model with HPCFE, where input=$\{\bm x^{i}, prev\}$, output=$y_i$.}
\State{Add $g_i(\{\bm x^{i+1},prev\})$ to $prev$.}
\EndFor
\State{Finally we cascades the $g_i$ HPCFE models where $i \in$ [2 to $M$] such that {$y^i$} is the output and  $[\bm{x},f_{i-1}(\bm{x}),f_{i-2}(\bm{x}),.....f_1(\bm{x})]$ are the input. The resulting expression of deep-HPCFE can be obtained as :
$f_i(\bm{x})=g_i([\bm{x},f_{i-1}(\bm{x}),f_{i-2}(\bm{x}),.....f_1(\bm{x})])$}
\Ensure{ We get output as $g_1$ HPCFE model and $g_i$ where i $\in$ [2 to M].}
\end{algorithmic}
\end{algorithm}

\begin{figure}
    \centering
    \includegraphics[scale=0.7]{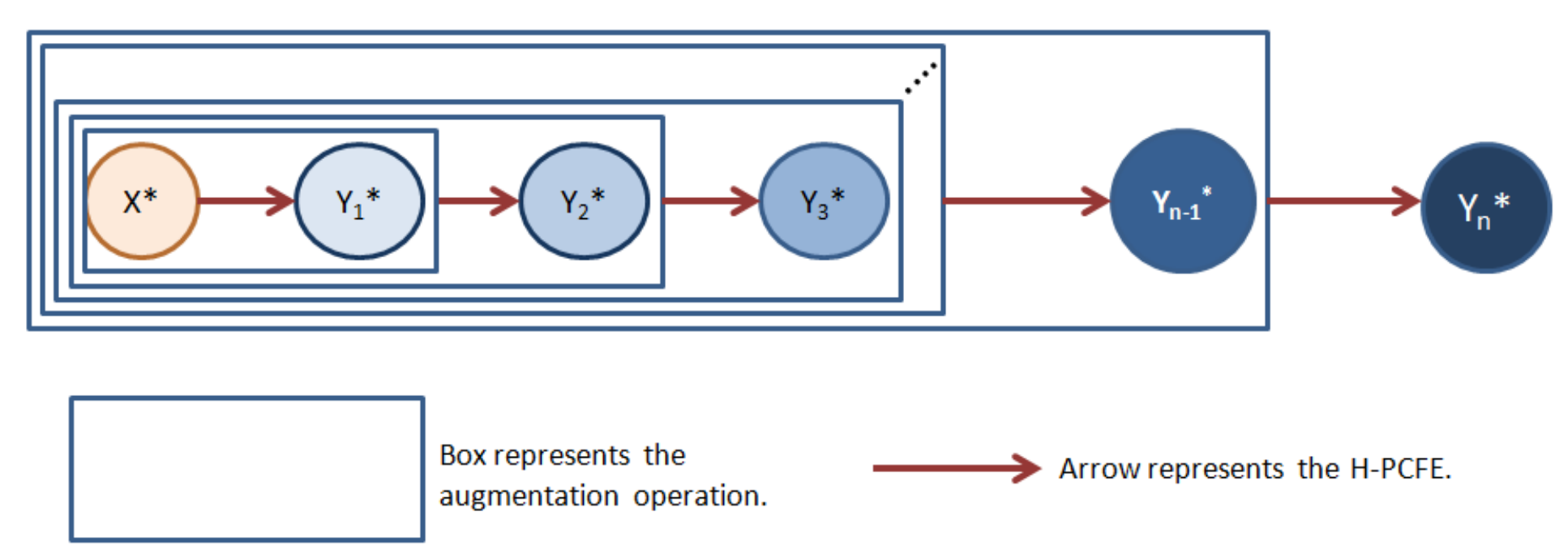}
    \caption{A schematic illustration of Multi-fidelity H-PCFE. Here ${X^{\ast}}$ represents the input data and ${Y_n^{\ast}}$ denotes the output data corresponding to the $n^{th}$ fidelity}
    \label{multi_HPCFE}
\end{figure}

\begin{algorithm}[ht!]
\caption{deep H-PCFE prediction}\label{alg:hpcfe_pred}
\textbf{Requirements:} Trained deep-HPCFE with $g_i$ HPCFE model  corresponding to the each fidelity level, where $i$ $\in$ [2 to M], and the input vector $\bm x^*$ at which output predictions are sought.
\begin{algorithmic}[1]
\State{Declare $C$ as a null matrix}

\For{$i \gets 1 $ to $ M \do$}
\State{We will find $\boldmath{y_i}$ i.e output at $i^{th}$ fidelity level by $\boldmath{y_i}$ = $g_i(\bm x^*,C)$ }
\State{Add previous fidelity level output to input such that C=(C,$g_i(\bm x^*)$)}
\EndFor
\Ensure{The highest fidelity ($n^{th}$ level) output prediction at given input $\bm x^*$ can be obtained as: \\
$f_M(\bm x^*)=g_M(\bm x^*,g_{M-1}(\bm x^*,g_{M-2}(\bm x^*).....g_1(\bm x^*))....g_3(\bm x^*,g_2(\bm x^*,g_1(x*)),g_1(\bm x^*)),
g_2(\bm x^*,g_1(x*)),g_1(\bm x^*))$}
\end{algorithmic}
\end{algorithm}
A schematic depiction of the proposed approach is shown in \autoref{multi_HPCFE}.
\section{Application of deep H-PCFE for uncertainty quantification}
\subsection{Context of the problem}
Let us consider a general computational model of the form $y=g(\bm{x})$ with $y$ as system response and $x$ as input parameters such that for a given realization of $x$, the computational model yields an output. When the input parameters vary in a certain range, accordingly, the output also varies. Here we aim to estimate the uncertainty in the output due to the variation in the input $\bm{x}$. The varying input can be represented as an N-dimensional vector of random variables $\bm{X}$, $\bm{X}=\left(X_{1},X_{2},....,X_{N}\right)$: $\Omega_{\bm X}\to \mathbb{R}^{N}$, with probability density function $P_{\bm X}(\bm x)$ and  having cumulative distribution function $F_{\bm{X}}\left(\bm{x}\right)=\mathbb P \left(\bm{X}\leq \bm{x}\right)$. Here $\mathbb P$ denotes the probability, and $\Omega_{\bm X}$ denotes the probability space. To compute the probability density of output $P_{\bm Y}(\bm y)$ we employ a surrogate model, $\hat{g}(\bm{x})$, which approximates the actual computational model through a mapping between the inputs and the output. Here the training samples are generated using design of experiments $\bm{\Xi}=\left[\bm X^{(1)},\bm X^{(2)},\ldots,\bm X^{(N_s)}\right]^T$ for the corresponding function evaluations $\bm Y = \left[Y^{(1)}, Y^{(2)}, \ldots, Y^{(N_s)}\right]^T$, with $N_s$ as the number of training samples. Mathematically it can be represented as:
\begin{equation}
g(\bm{x})=\hat{g}(\bm{x};\bm{\theta})+\epsilon,
\end{equation}
where $\bm{\theta}$ denotes the set of surrogate parameters and $\epsilon$ denotes the surrogate error. As we discussed earlier, most of the time, a simple surrogate fails to handle multi-fidelity data. Suppose we have pair of  input parameters and corresponding responses of the system from multiple sources having $M$ different fidelities (sensors with different precision) of the form $\{\mathbf X^{1},\bm y^{1}\}$, $\{\mathbf X^{2},\bm y^{2}\}$,$......\{\mathbf  X^{M},\bm y^{M}\}$, the challenging task here will be to fuse the multi fidelity data. To that end, we leverage the proposed deep H-PCFE to integrate the information efficiently from multi-fidelity data.

\subsection{Numerical examples}
Three numerical examples are presented in this section to demonstrate the application of the proposed Deep-H-PCFE for multi-fidelity modelling in uncertainty quantification. All of the examples employ either analytical or numerical methods to generate multi-level data. Despite the fact that any other experimental design might be employed instead, uniformly distributed sample points are used as training data here. The goal here is to quantify the uncertainty in the output response in terms of the probability density function of the quantity of interest. However, it is also required to evaluate the efficacy of the proposed method. For that, we compare the results obtained by the proposed framework to that of Crude MCS results. A detailed illustration is provided in each numerical example.

\subsection{Benchmark mathematical examples}
\subsubsection{A Pedagogical example}
\begin{figure}[ht!]
    \centering
    \includegraphics[scale=0.3]{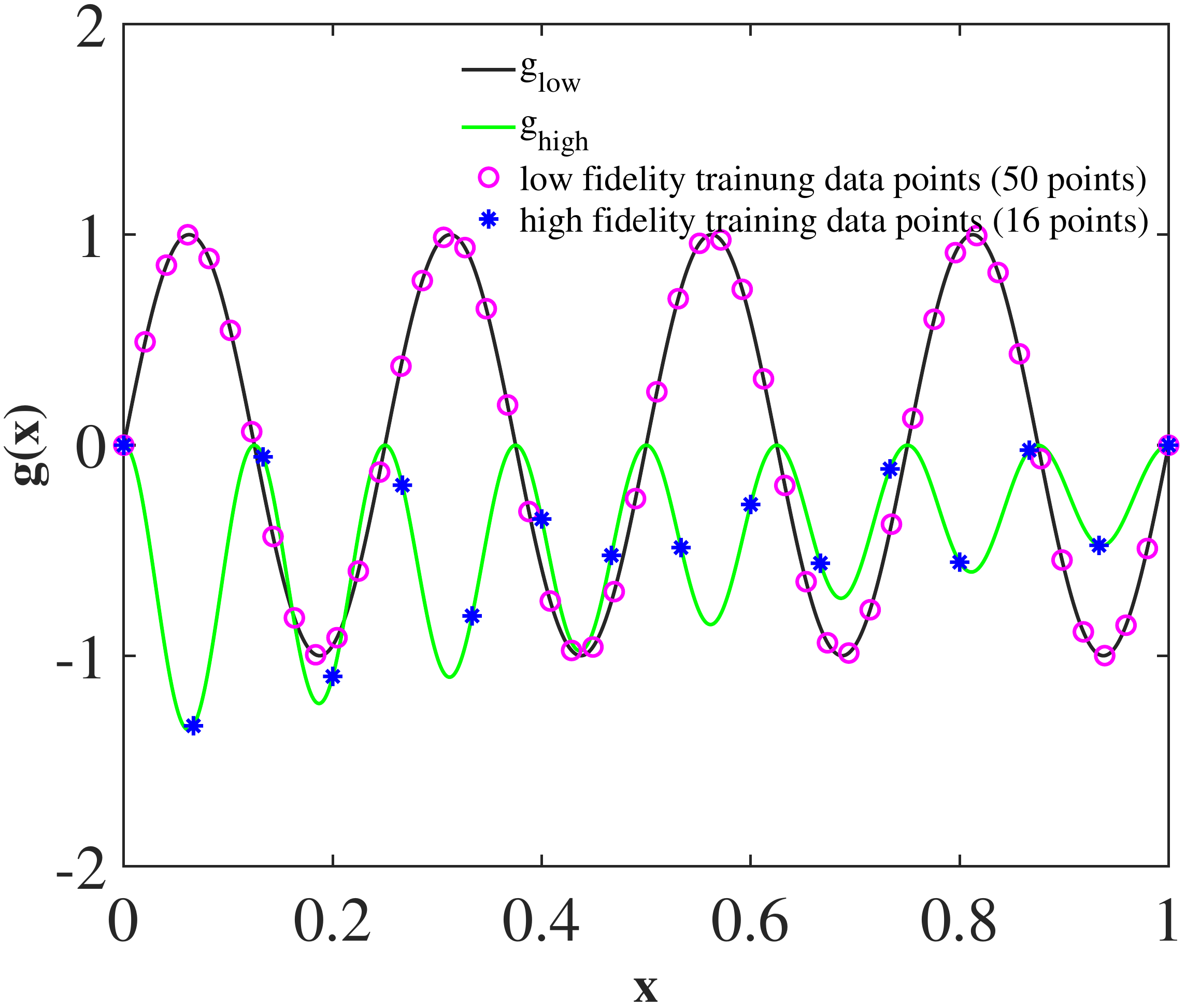}
    \caption{Exact low-fidelity and high-fidelity functions (refer to \autoref{hlf}), and the chosen data points for training the MF-HPCFE (50 low-fidelity points and 16 high-fidelity points)}
    \label{datagen_pdf}
\end{figure}

We start with a relatively simpler example, where we deal with data of two levels of fidelities, i.e. high fidelity and low fidelity. While the low fidelity function is a sinusoidal wave with four periods, the  high fidelity function is obtained by transforming low fidelity non-uniform scaling and quadratic non-linearity. The expressions of analytical formulations used for generating high-fidelity and low-fidelity data are given below in Eq. \eqref{hlf}:

\begin{equation}\label{hlf}
\begin{gathered}
g_{1ow}(x)=\sin (8 \pi x) \\
g_{\mathrm{high}}(x)=(x-\sqrt{2}) g_{1ow}^{2}(x).
\end{gathered}
\end{equation}
Let us suppose we have a finite number of low-fidelity data and very few high-fidelity data points. The primary objective here is to achieve the closest approximation for the high-fidelity response data. In general, the efficiency of the surrogate is determined by the number of training samples and the accuracy metrics. However, here the constraint is more challenging as we need to achieve accurate prediction from the limited number of low-fidelity data and fewer higher-fidelity points. both low-fidelity and high-fidelity data are used. For the current example, the number of low-fidelity points used to train the model is n1=50, whereas the number of high-fidelity points utilized is n2=16. The points are chosen such that $n_1$ points form data set $A_1$ and $n_2$ points form data set $A_2$, where $A2 \subseteq A1$ \cite{article_multi}. We provide a further visual illustration of generated samples in \autoref{datagen_pdf}.  
The vanilla H-PCFE trained solely with high-fidelity data leads to inaccurate prediction as the data is too scarce to identify the underlying signal. 

\begin{figure}[ht!]
    \centering
    \subfigure[]{
    \centering
    \includegraphics[width=.4\textwidth]{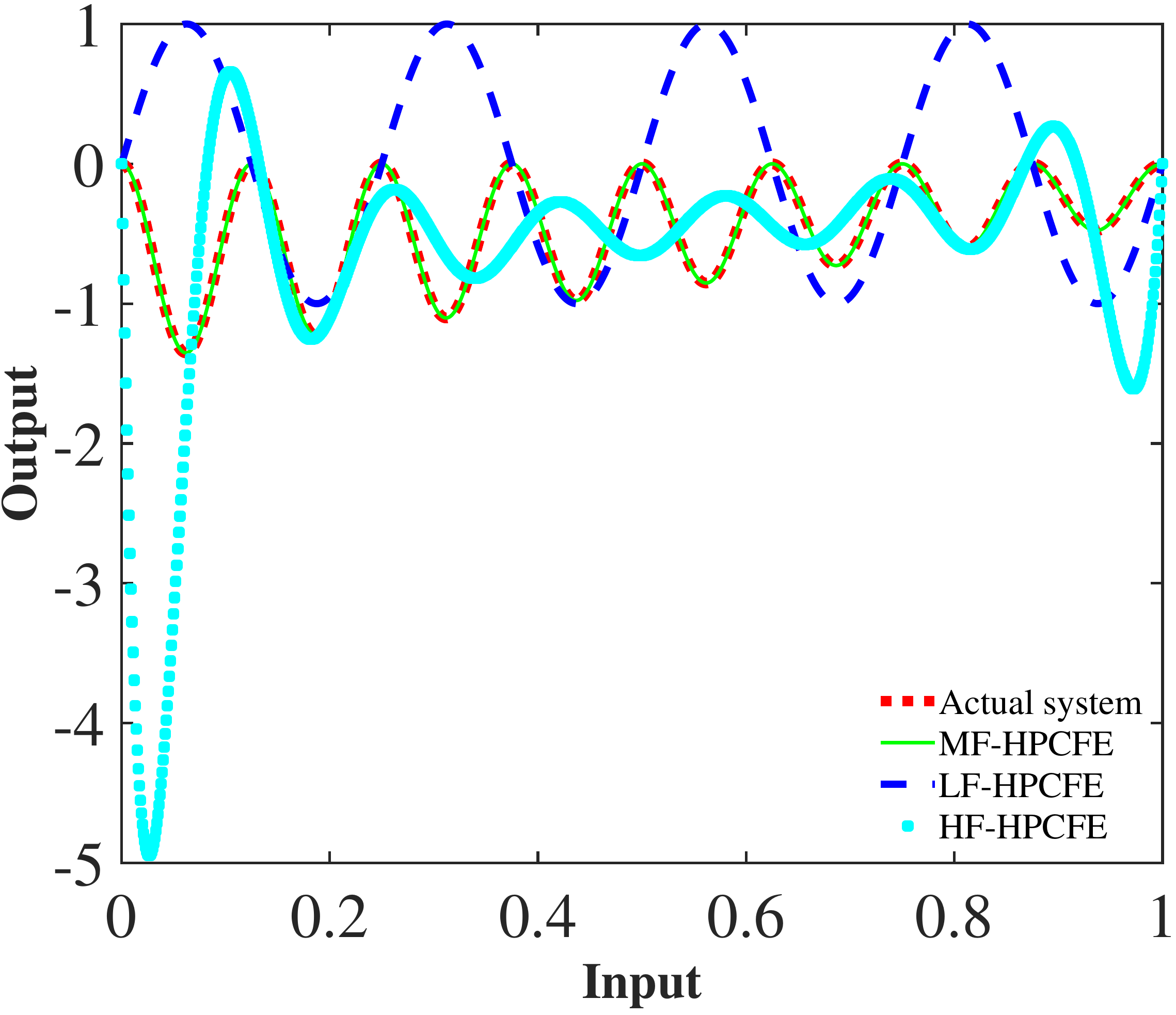}}
    \subfigure[]{
    \centering   
    \label{}
    \includegraphics[width=.4\textwidth]{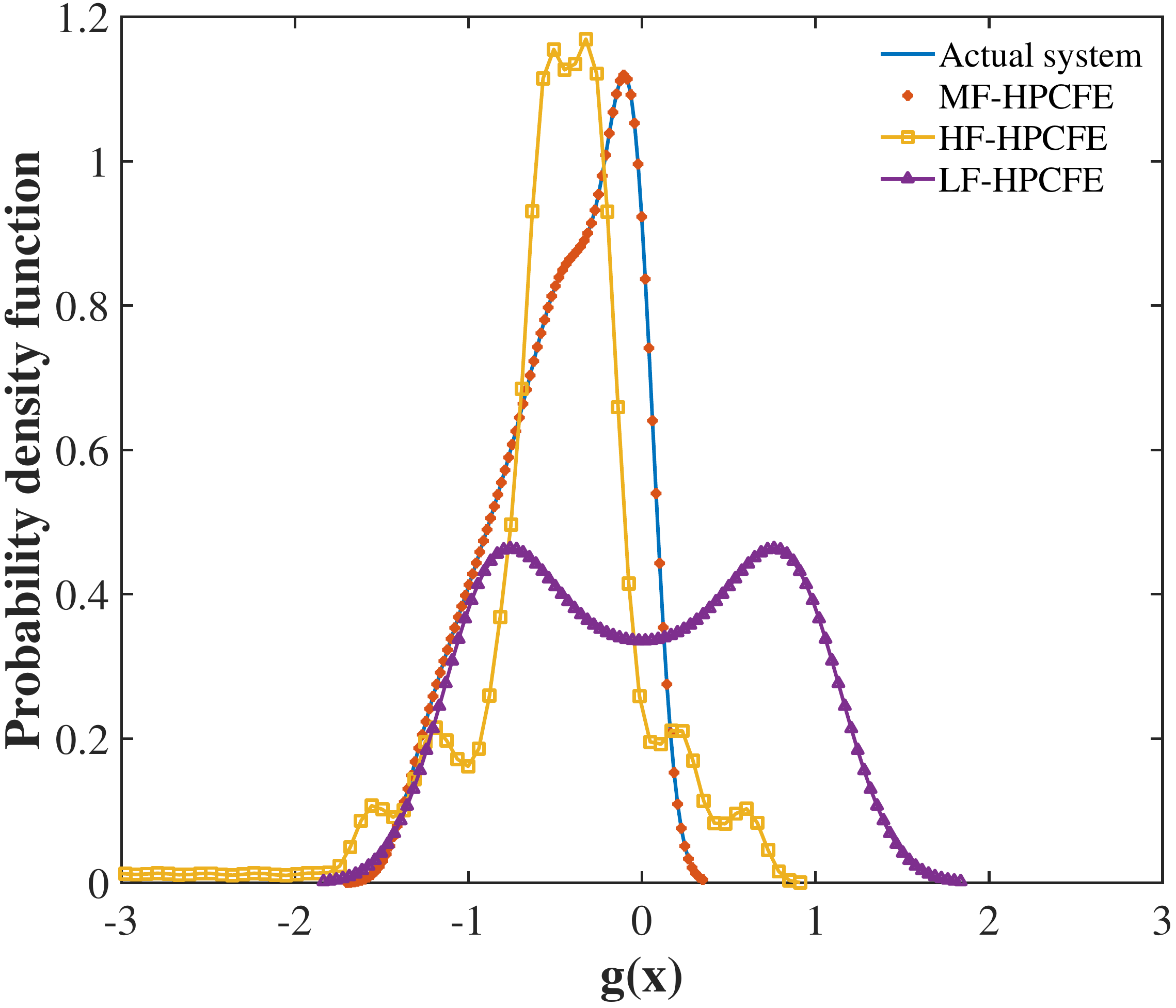}}
    \caption{Predictions of deep H-PCFE (a) predicted function by deep-HPCFE in comparison with LF-HPCFE, HF-HPCFE and ground truth function (b) Probability density function corresponding predictions and ground truth values.}\label{ypred_prob1}

\end{figure}

The prediction results are demonstrated in the \autoref{ypred_prob1}. The \autoref{ypred_prob1}(a) showcases high fidelity function  predicted by deep H-PCFE in comparison with predictions of the H-PCFE models trained with only high fidelity data and low fidelity data, and actual high fidelity function. Further, to evaluate the performance of the proposed framework in uncertainty quantification, we examine the probability density plots depicted in \autoref{ypred_prob1}(b). In the depicted results, the deep H-PCFE is referred to as multi-fidelity H-PCFE (MF-H-PCFE). The results elucidate that the prediction by MF-HPCFE matches closely with the ground truth for the prescribed number of high-fidelity and low-fidelity training data points. Similarly, the apparent visual resemblance can be seen in probability density plots of MF-H-PCFE predictions and the ground truth values, where x follows a uniform distribution. On the other hand, the predictions of H-PCFEs, trained either solely with the high-fidelity data or solely with the low-fidelity data, referred to here as HF-HPCFE and LF-HPCFE, deviate from the ground truth values significantly.

\begin{figure}[ht!]
    \centering
    \includegraphics[width=.4\textwidth]{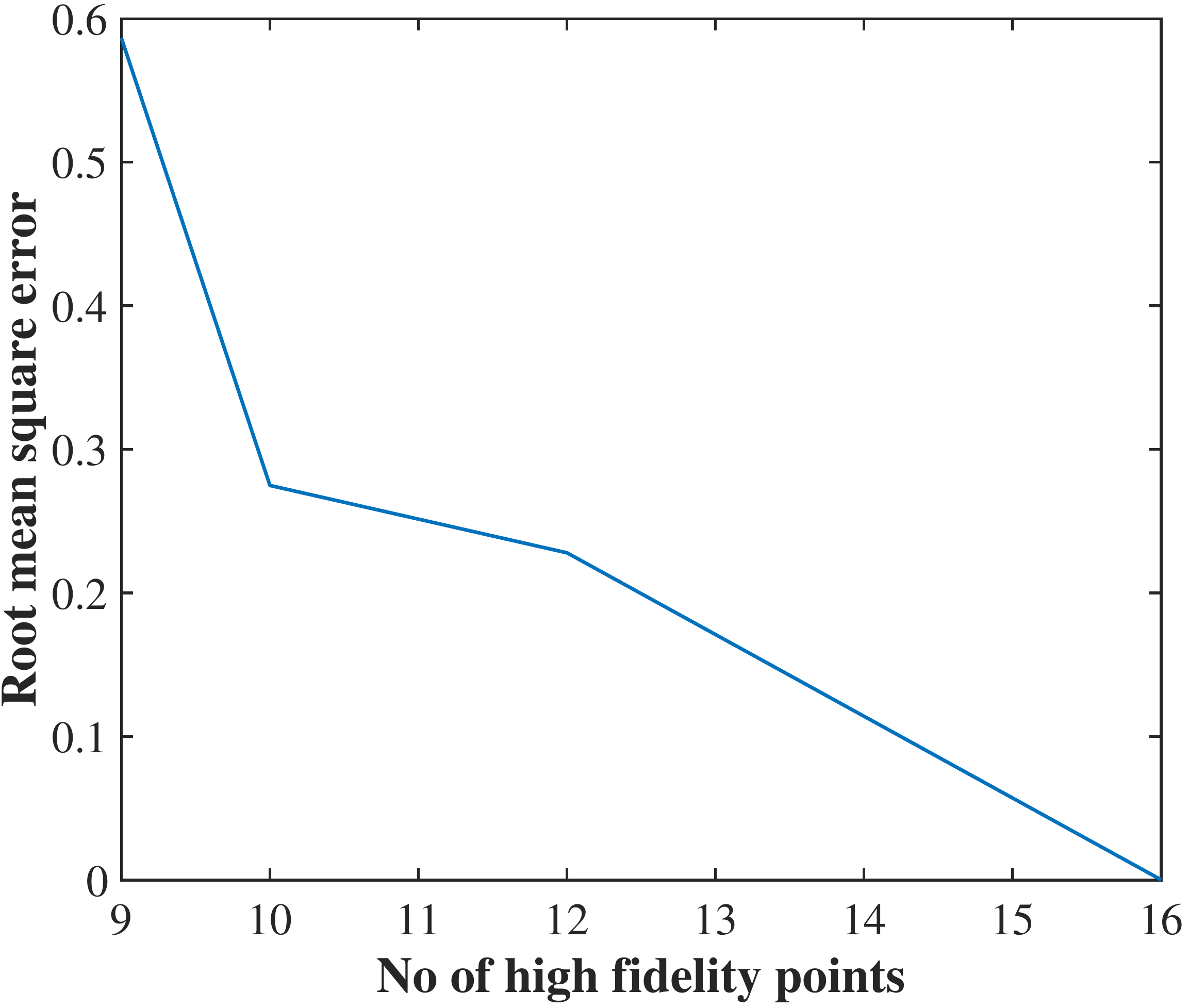}
    \caption{Convergence of the prediction error ( Root mean square error of actual and predicted values) with increase in number of high-fidelity points, keeping number of low fidelity points =50}
    \label{prob1_error_plot}
\end{figure}

In order to further investigate the effect of the number of high-fidelity data points on the accuracy of the model prediction, a case study has been carried out and is presented in \autoref{prob1_error_plot}. The illustrated plot shows the variation of the root mean square error between predictions
by MF-HPCFE and ground truth with a number of high-fidelity data points. The plot clearly infers that with an increase in the number of high-fidelity points, root means square error decreases, and the model enhances the performance as expected.

\subsubsection{Buckling of plate problem}\label{buckling_pl}
As the second example, we illustrate the problem of buckling of a plate. Contrary to the first numerical example, the problem we consider here is a more challenging problem and corresponds to a practical
application setting. The boundary condition is set to be simply supported from all sides, whereas the uniform load is applied along the length. For the given problem setting, the primary objective is to quantify the uncertainty in the critical buckling load of the plate. To estimate the critical buckling load, the obvious choice is to rely on the analytical formulations \cite{THAI20138310}. However, these formulation does not yield accurate results due to the underlying assumptions. More realistic results can be obtained   
through finite element computational models. While the analytical formulations are considered as the low-fidelity model, the high fidelity and the intermediate level fidelity data are generated by employing finite element simulations with 3-D elements (solid section element) and 2-D elements (shell section element). Considering the influence of various factors, plate dimensions and material properties are taken as random variables. Thus the random input variables to the mathematical model/computational model include
 length (a), breadth (b), thickness (t), Young's modulus(E), and Poisson's ratio($\mu$). The detailed information on these parameters is presented in the \autoref{buckling_table}.
\begin{table}[H]
    \centering
    \caption{Distribution type and distribution of input parameters used for simulating critical buckling load of the plate.}
    \label{buckling_table}
\begin{tabular}{lcccc} 
\hline
{Variable no.} & {Variable} & {Mean} & {COV} & {Distribution} \\
\hline
1 & $a$ {m} & $3$  & 0.05 & Normal \\
2 & $b$ {m} & $2$  & 0.05 & Normal \\
3 & $t$ {m} & $0.2$  & 0.075 & Raileigh \\
4 & $E$ {kpascal} & $2\times10^{9}$  & 0.1 & Lognormal \\
5 & $\mu$  & $0.3$  & 0.025 & Lognormal \\
\hline 
\end{tabular}
\end{table}
 
\begin{figure}[H]
    \centering
    \subfigure[]{
    \includegraphics[scale=.3]{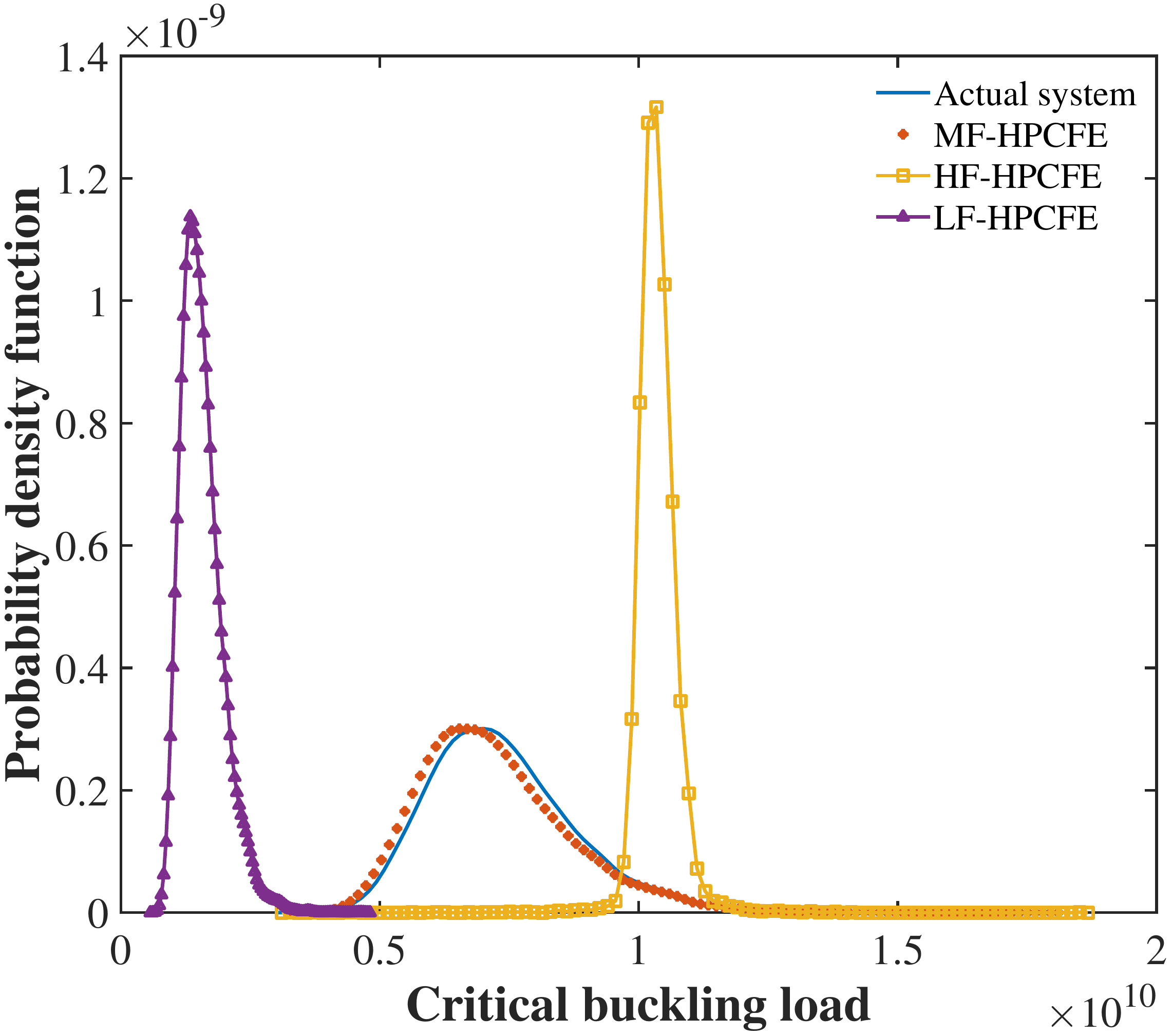}}
    \subfigure[]{
    \includegraphics[scale=0.3]{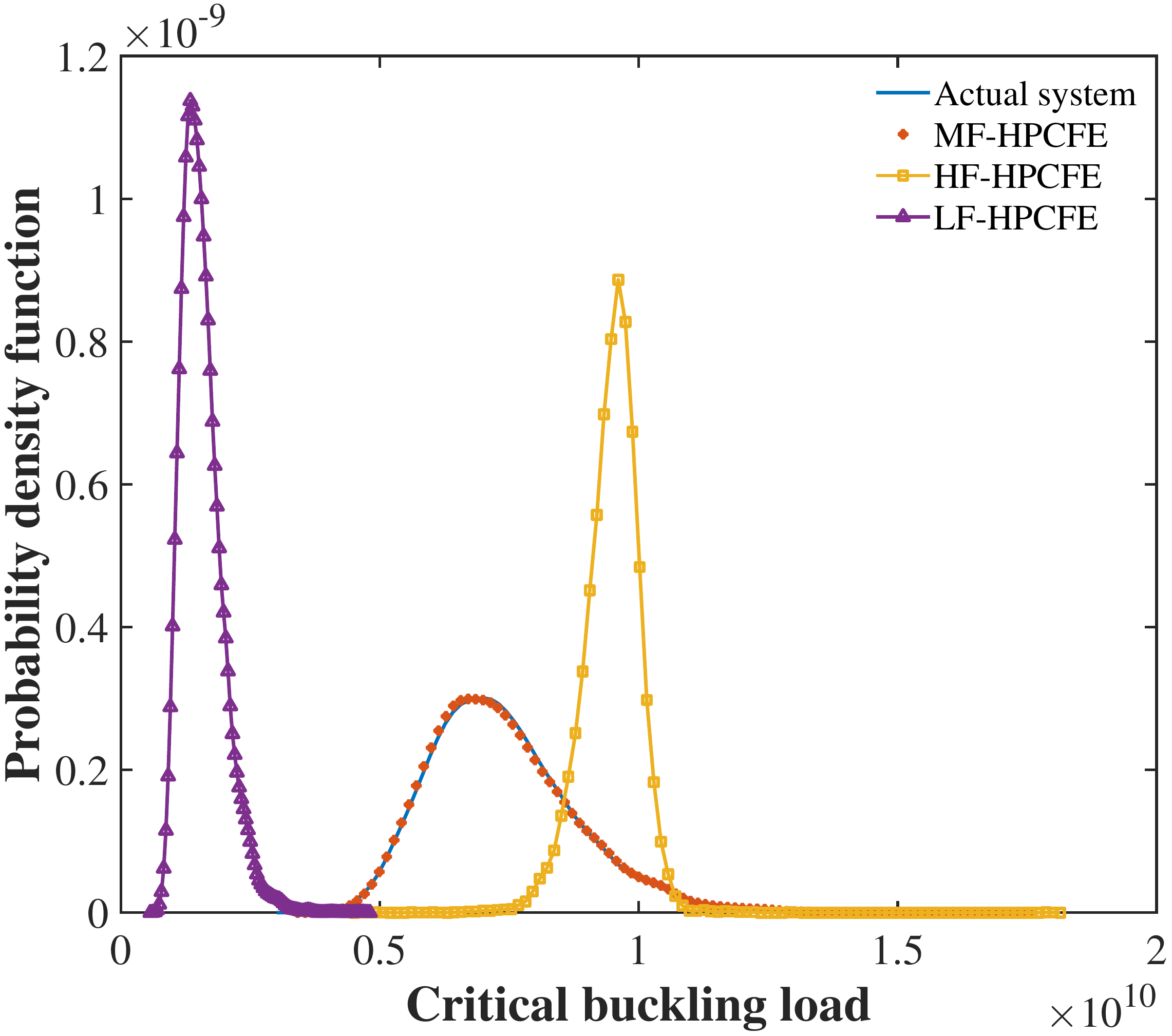}}
    \caption{Probability density function corresponding to the predictions of (a) deep-HPCFE trained with 210 LF points, 15 $HF_1$ points, and 5 $HF-2$ points  and (b) deep-HPCFE trained with 210 LF points, 15 HF1 points, 8 HF2 points, where LF denotes low-fidelity, $HF_1$ denotes intermediate fidelity, and $HF_2$ denotes high-fidelity}\label{Buckling of plates1}
\end{figure}

 The results of the study are demonstrated in \autoref{Buckling of plates1}.  
The PDF plots of the output (critical buckling load) are presented in \autoref{Buckling of plates1}. The reported results investigate the efficacy of the proposed method for two sets of samples having different numbers of high fidelity points while keeping the low fidelity points ($LF = 210$) and intermediate fidelity sample points ($HF1 = 15$) constant. The PDF plots of the deep-HPCFE (MF-HPCFE) predictions show good agreement with that of the ground truth. As is seen in \autoref{Buckling of plates1}(a), though the results of MF-HPCFE deviate slightly from the PDF of the true values, with an increase in the number of high fidelity points to ($HF2=5$) the PDF plots of the MF-HPCFE emulates the PDF plots of the ground truth almost exactly. The noteworthy observation is that similar to the previous example, the surrogate models trained with single fidelity data, HF-HPCFE and LF-HPCFE, fail to obtain the desired performance.

\section{Application of deep H-PCFE for the enhanced digital twin framework}
\subsection{Context of the problem}
A DT describes the time evolution of a physical system from the nominal model to the updated state of the system based on the response measurements achieved through data acquisition systems. For engineering dynamical systems, the nominal model is generally a physics-based model which has been verified, validated, and calibrated. We explain the essential ideas through a single degree of freedom (SDOF) dynamic system. We begin with considering the equation of motion of a single degree of freedom dynamic system \cite{inman1994engineering} is expressed of the form:
\begin{equation}\label{sdof_eq1}
m_0 \ddot{u}_0(t){t} +c_0 \dot{u}_0(t){t} + k_0 u_0(t) = f_0(t)
\end{equation}
We call the system given by Eq. \eqref{sdof_eq1} as the nominal dynamical system. Here $m_0$, $c_0$, and $k_0$ are the nominal mass, damping, and stiffness coefficients. The forcing function and the dynamic response are denoted by $f_0(t)$ and $u_0(t)$, respectively. It is worthwhile to note that the described SDOF model in Eq. \eqref{sdof_eq1} can be regarded as either a simplified model of a more complex dynamic system or the representation of the dynamics of a modal coordinate of a multiple-degree freedom system. 
We proceed further by diving $m_0$ throughout the equation, and thus the equation of motion Eq. \eqref{sdof_eq1} can be rewritten as:
\begin{equation}\label{sdof_eq2}
{\ddot u}_0(t) + 2 \zeta_0 \omega_0 {\dot u}_0(t) +  \omega^2_0  u_0(t) = {\frac { f(t) } {m_0} }
\end{equation} 
Here the undamped natural frequency ($\omega_0$), the damping factor ($\zeta_0$) and the natural time period of the underlying undamped system are expressed as:
\begin{align}
    \begin{split}\label{eq:a}
      \omega_0 & =\sqrt{\frac {k_0} {m_0}}, 
    \end{split}\\
    \begin{split}\label{eq:b}
      {\frac {c_0} {m_0} } & = 2 \zeta_0  \omega_0  \quad \text{or} \quad  \zeta_0 = {\frac {c_0} {2\sqrt{k_0 m_0} } } 
    \end{split}\\
    \begin{split}
       T_0 &= \frac {2\pi} {\omega_0} 
     \end{split}
\end{align}
Taking the Laplace transform of Eq. \eqref{sdof_eq2}, the expression results in:  
\begin{equation}\label{sdof_eq3}
s^2  U_0(s) + s 2 \zeta_0  \omega_0 U_0(s) + \omega_0^2 U_0(s) = {\frac {  F_0(s)} {m_0} },
\end{equation}
where $U_0(s)$ and $F_0(s)$ are the Laplace transforms of $u_0(t)$ and $f_0(t)$ respectively. Solving the equation associated with coefficient of $U_0(s)$ in Eq. \eqref{sdof_eq2} without the forcing term, the complex natural frequencies of the system are given by:
\begin{equation}\label{freqs}
\lambda_{0_{1,2}}= - \zeta_0  \omega_0 \pm {i} \omega_0 \sqrt{1-\zeta_0^2}= - \zeta_0  \omega_0 \pm {i} \omega_{d_0}  
\end{equation}
Here the imaginary number $i=\sqrt{-1}$ and the damped natural frequency is expressed as
\begin{equation}\label{damped_freq}
\omega_{d_0} = \omega_0 \sqrt{1-\zeta_0^2}
\end{equation}
For a damped oscillator, at resonance, the frequency of oscillation is given by $ \omega_{d_0} < \omega_0$. Therefore, for positive damping, the resonance frequency of a damped system is always lower than the corresponding underlying undamped system. 

\begin{figure}[ht!]
	\centering
	\includegraphics[width=0.8\textwidth]{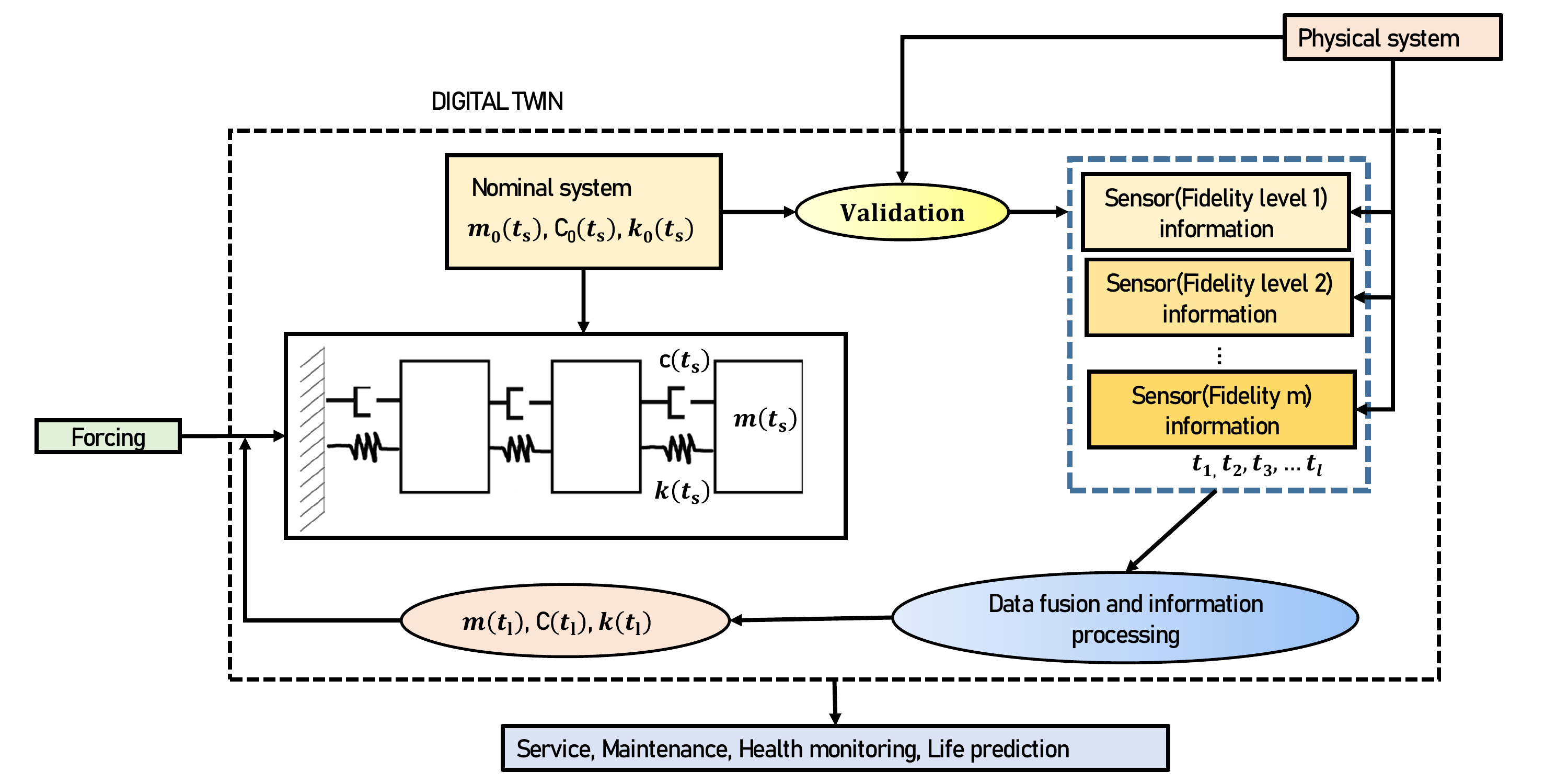}
	\caption{The overview of constructing a digital twin for a multi-degree-of-freedom dynamic system.}\label{DT_scheme} 
\end{figure}

\begin{figure}[ht!]
	\centering
	\includegraphics[width= 0.4\textwidth]{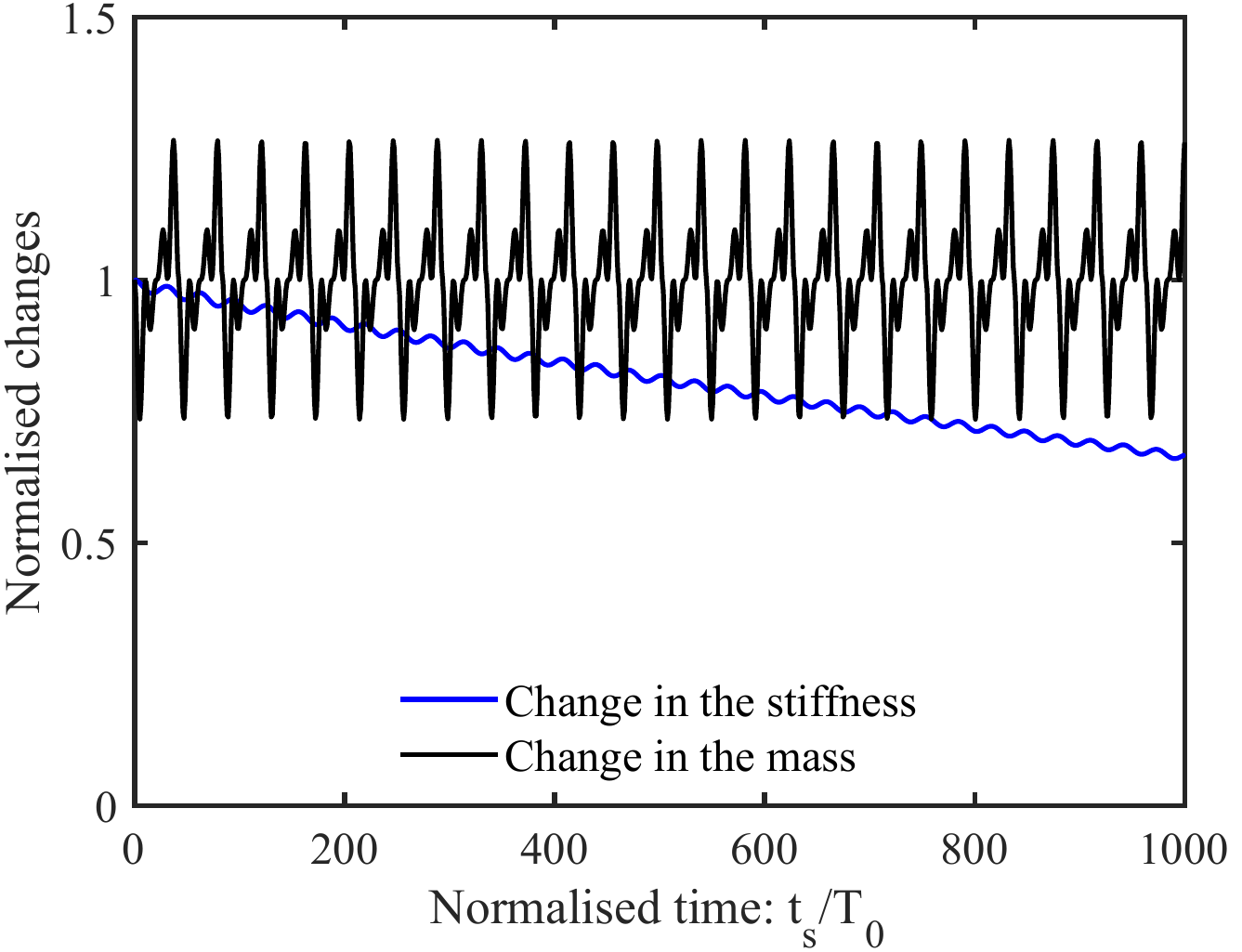}
	\caption{Examples (refer to \autoref{var_fnmK}) of model functions representing long-term variation in the mass and stiffness properties of a digital twin system.}\label{property_changes}
\end{figure}

Now for the same single degree of freedom system, the  digital twin model can be described as:
\begin{equation}\label{sdof_dt}
m(t_s) \ddot{u}(t,t_s){t} + c(t_s) \dot{u}(t,t_s){t} + k(t_s) u(t,t_s)=f(t,t_s)
\end{equation}
Here $t$ represents the system time $t_s$ represents the "slow time" (service time), which can be considered as a time variable having a much slower variation than $t$. When the system degrades over the service period, the system properties, mass $m(t_s)$, damping $c(t_s)$, stiffness $k(t_s)$ and forcing $F(t, t_s)$ change with $t_s$ also varies in the stipulated time. Moreover, the forcing term is a function of time $t$ and slow time $t_s$. Thus, in essence, Eq. \eqref{sdof_dt} represents the mathematical model of a digital twin of a (Single degree of Freedom) SDOF dynamical system having the response $u(t,t_s)$ such that when $t_s=0$, that is at the beginning of the service life, the model represents the nominal model \autoref{sdof_eq1}. 
In a realistic scenario, the sensors, along with the data acquisition systems, collect measurements at discretized time instances of the time $t_S$. Upon obtaining the sensor data, the unknown functional form of the system properties and the forcing term with time $t_s$ can be estimated. A general overview representing the implementation of a digital twin for a single-degree-of-freedom dynamic system is schematically depicted in \autoref{DT_scheme}.
While the digital twin model enables to obtain the embodied  functions of $k(t_s)$, $m(t_s)$, and $c(t_s)$ effectively, it is apparent that the feasible choices of the functional forms are not limited but several. For instance, the stiffness function $k(t_s)$ can be considered as a random function (i.e., a random process). Further we also assume that damping is small so that the effect of variations in $c(t_s)$ is negligible. In effect only variations in the mass and stiffness is considered. Thus it is legitimate to express functional forms $k(t_s)$ and $m(t_s)$ as follows:
\begin{equation}\label{cond2}
\begin{split}
& k(t_s) = k_0(1+\Delta_k(t_s)), \quad \text{and} \\
& m(t_s) = m_0(1+\Delta_m(t_s)),
\end{split}
\end{equation}  
where it is given that $\Delta_k(t_s)=\Delta_m(t_s)=0$ for $t_s=0$. Typically, the stiffness of the system $k(t_s)$ deteriorates with the progression of time, and thus it is assumed to be a decaying function over a long time. However, the same is not the case with $m(t_s)$. Loading of cargo and passengers and use of fuel with the progression of flight exemplify such a situation where $m(t_s)$ can be an increasing or decreasing function. In light of the above discussions, the following  functions are chosen conveniently as representative examples:
\begin{equation}\label{var_fnmK}
\Delta_k(t_s) = e^{-\alpha_k t_s} {\frac{(1+ \epsilon_k \cos(\beta_k t_s))} {(1+\epsilon_k)} }-1 
\end{equation}
\begin{equation}\label{var_fnm}
\Delta_m(t_s) = \epsilon_m \text{ SawTooth} (\beta_m ( t_s - \pi/\beta_m) ) \sin^2(2 \beta_m t_s )
\end{equation}
Here the function, SawTooth$(\bullet)$, denotes a sawtooth wave with a period $2\pi$. A visualization of the overall variation of the modeled stiffness and mass of the system is provided in \autoref{property_changes}, where time is normalized to the natural time period of the nominal model.
The parameter values in the functions are chosen to be $\alpha_k=4\times 10^{-4}$,  $\epsilon_k=0.05$,  $\beta_k=0.2$,  $\beta_m=0.15$ and  $\epsilon_m=0.35$. One may consider fatigue crack growth in an aircraft over repeated pressurization as a valid reason for stiffness degradation over time in a periodic manner. On the other hand, the mass of the aircraft may increase and decrease over the nominal value due to re-fuelling and fuel burn over a flight period. Ideally, the digital model aims to track the variation in the system properties through the measured sensor data. 

In essence, the digital twin model corresponding to the single degree of freedom system enables us to track the variation of the mass and stiffness ( $\Delta m$, $\Delta k$ ) with time. To that end, the real-time measurements of displacements and the dampened natural frequency of the system are inevitable. However, in practice, several factors impede the seamless acquisition of sensor data, and thus the collected data may have different fidelities. 
In the context of digital twin having two fidelities in the measured data, i.e, a high fidelity data $\mathcal D_f =  \left\{\bm x_i, \bm y_{f, i} \right\}_{i=1^{N_f}}$ and a low-fidelity data $\mathcal D_c = \left\{\bm x_i, \bm y_{c, i} \right\}_{i=1^{N_c}}$, we aim to obtain the mapping $\mathcal M: \bm x \mapsto y_f$. Here input $\bm x$ represents the time,  whereas the output $\bm y$ represents system parameters (e.g., mass, stiffness, damping). 

\subsection{Numerical example}
By definition, the digital twin model solves an inverse problem. In a real-time setting, we lack first-hand information on the varying system parameters ($\Delta m$ or $\Delta k$ ), whereas the digital twin model learns the parameters from the measured data. However, we note that, for the sake of numerical illustration and validation of the proposed framework in the present work, we utilize synthetic data instead of the actual field data. The synthetic data is generated by assuming some stiffness and mass degradation functions (see Eq. \eqref{var_fnmK}) to be true for high-fidelity data. In order to obtain low-fidelity data, we perform certain operations on the high-fidelity data function. The modified stiffness degradation function used here to generate the low-fidelity data is as follows:   
\begin{equation}
\label{k_fun}
{\Delta_{low_k}}=0.75{\Delta _k} + 0.01\sin (1000 + \frac{\pi }{{10}}t{\Delta _k}).
\end{equation}
For the low-fidelity mass degradation function, we have explored two plausible forms of equations, and are expressed as:
\begin{equation}
    {\Delta_{low_m1}}=0.75{\Delta_m}+0.01\cos(t\frac{\pi}{10})+0.025
    \label{m_fun}
\end{equation}
\begin{equation}
    {\Delta_{low_m2}}=0.25\text{SawTooth}(0.15(t - 20\frac{\pi }{3}))
    \label{m_fun_GPDT}
\end{equation}
\subsubsection{Digital twin using time domain data} \label{time_domain}
For the current study, we examine two cases in which the system properties of the dynamic system vary: in the first case, we assume that the change in the natural frequency with time is  solely due to the change in the mass of the system, where the stiffness and damping of the nominal model are kept unchanged, and in the second case, we consider that the time-evolution of the natural frequency is only due to the stiffness deterioration, where the mass and the damping of the nominal model are presumed to be invariants. We begin the numerical illustration by  considering the first case where the proposed framework is employed to track the  change in the mass of the SDOF system. The equation of motion of the digital twin of an SDOF system for a fixed value of $t_s$  with variation only in the mass property is given by:
\begin{equation}\label{sdof_eqm}
m_0(1+\Delta_m(t_s)) \Ddot{u}(t){t} +c_0 \dot{u}(t){t} + k_0 u(t) = f(t) .
\end{equation}
% Dividing with $m_0$ and solving the characteristic equation, the damped natural eigen frequencies can be obtained as:  
% \begin{equation}\label{eig_sol}
% \lambda_{s_{1,2}}(t_s)= - {\frac{\zeta_0 \omega_0}{1+\Delta_m(t_s)}} \pm {i} 
% {\frac{\omega_0 \sqrt{1+\Delta_m(t_s)-\zeta_0^2}} {1+\Delta_m(t_s)} }.
% \end{equation}
% Rewriting the \autoref{eig_sol}, in terms of natural frequency ($\omega_s(t_s)$), damping factor ($\zeta_s(t_s)$), and damped natural frequency ($\omega_{d_s}(t_s)$) of the digital twin, yields the following expression  
% \begin{equation}
% \lambda_{s_{1,2}}(t_s)= -\omega_s(t_s)\zeta_s(t_s)  \pm {i}  \omega_{d_s}(t_s),
% \end{equation}
% where the $\omega_s(t_s)$, $\zeta_s(t_s)$,and $\omega_{d_s}(t_s)$ are expressed as:  
% \begin{align}
% \omega_s(t_s) & =\omega_0 / \sqrt{1+\Delta_m(t_s)} \\
% \zeta_s(t_s) & =\zeta_0/\sqrt{1+\Delta_m(t_s)} \\
% \text{and} \quad
% \omega_{d_s}(t_s) & =\omega_s(t_s)\sqrt{1-\zeta^2_s(t_s)}
% ={\frac{\omega_0 \sqrt{1+\Delta_m(t_s)-\zeta_0^2}} {1+\Delta_m(t_s)} }.
% \end{align}
% Here we note that these three fundamental properties of the system ($\omega_s(t_s)$, $\zeta_s(t_s)$,and $\omega_{d_s}(t_s)$) change with the slow time $t_s$.

\begin{figure}[ht!]
	\centering
	\subfigure[]
	{\includegraphics[width=0.45\textwidth]{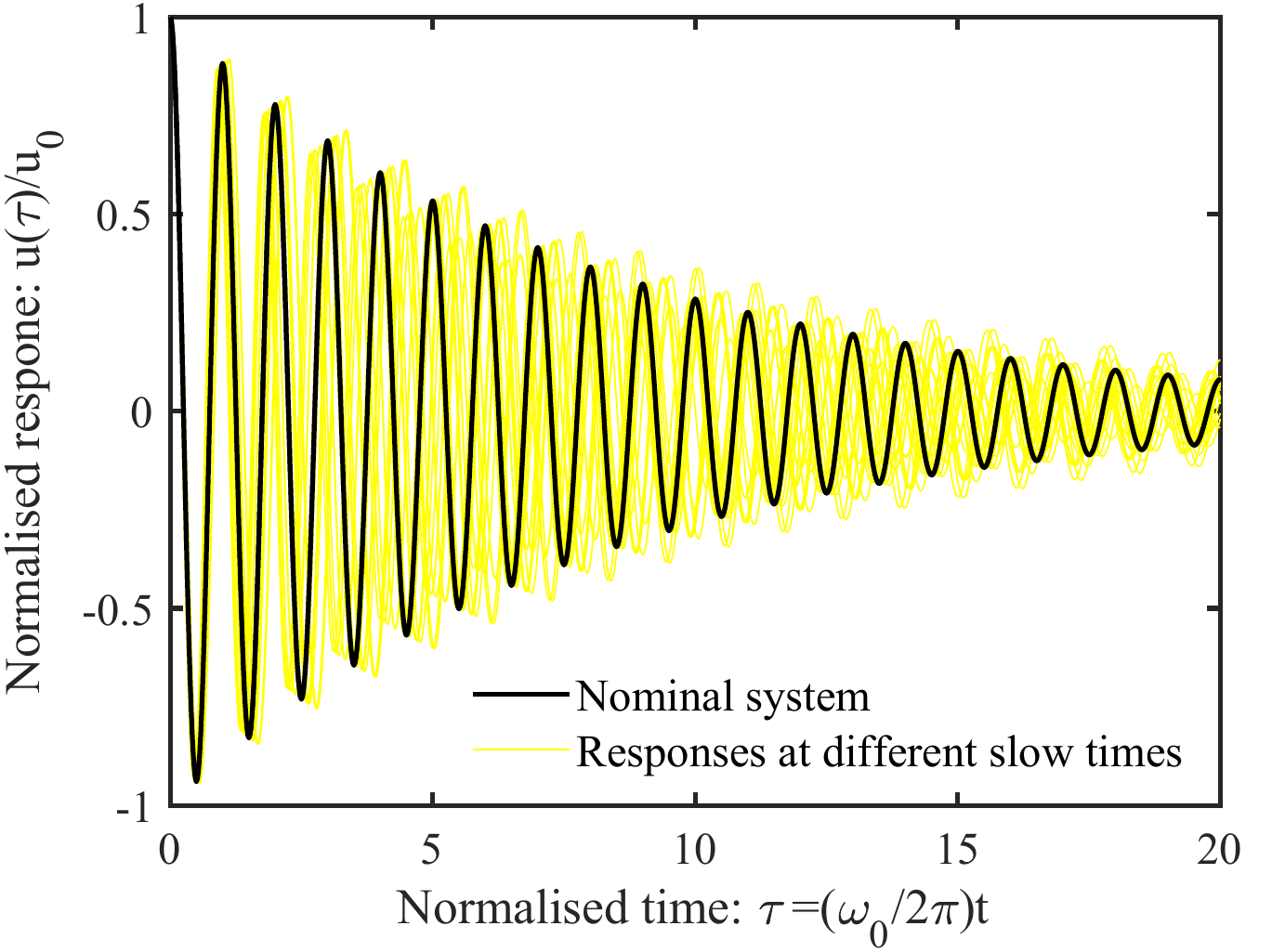}} 
	\subfigure[]
	{\includegraphics[width=0.45\textwidth]{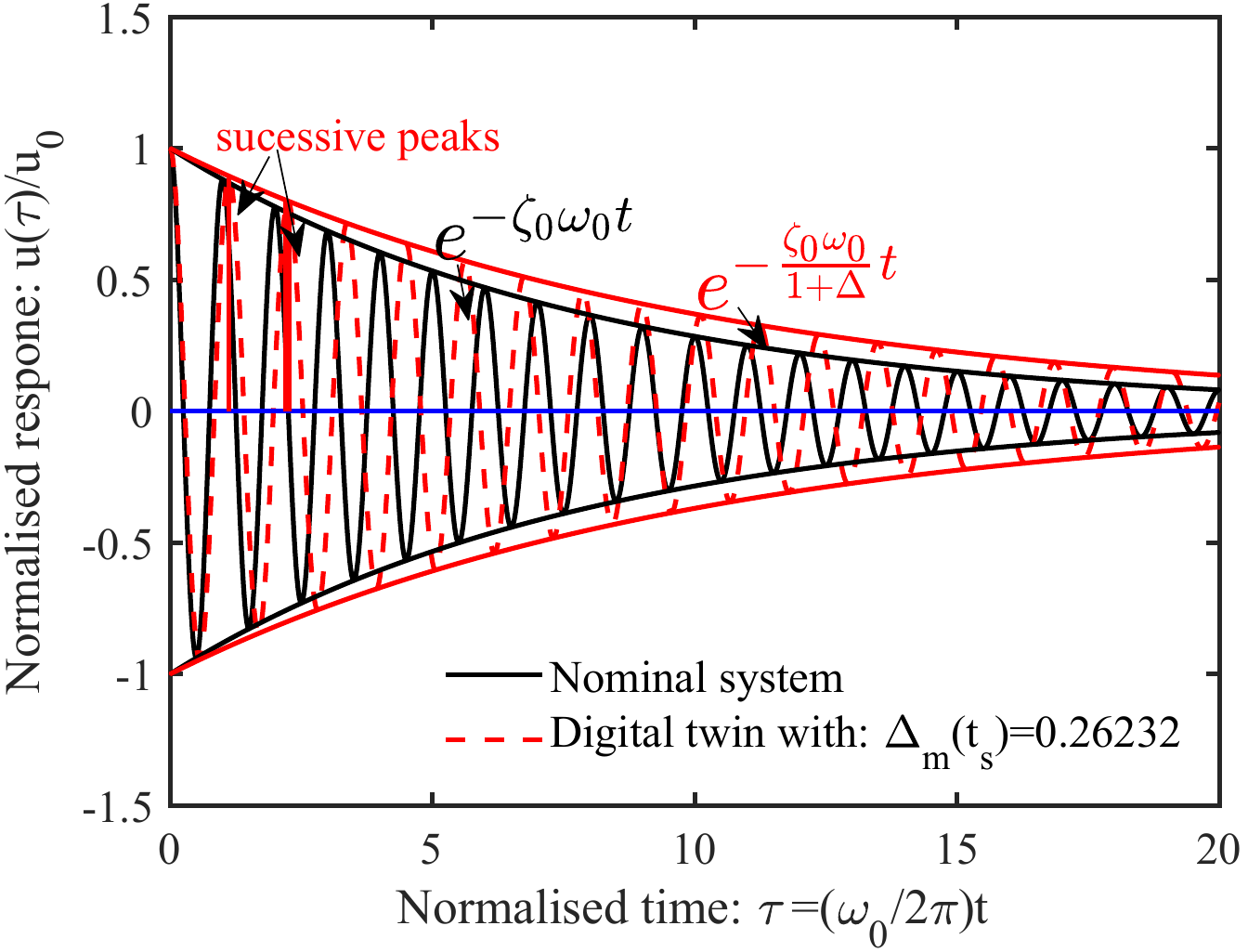}}
	\caption{Normalised response of the oscillator in the time domain due to an initial displacement plotted as a function of normalized time $\tau = t/T_0$ and damping factor $\zeta_0=0.02$.(a) The nominal system and measurements at different slow times(b) Analysis of the response due to an initial displacement.31}\label{timed_response}
\end{figure}

The free vibration response of a dynamic system is one of the most simplest and convenient measurements that can be used to develop the digital twin. The dynamic response of the digital twin model described in Eq. \eqref{sdof_eqm} due to an initial displacement $u_{i_0}$ and initial velocity $\dot u_{i_0}$ can be expressed as following \cite{inman1994engineering}: 
\begin{align}\label{damped_resp}
u_s (t_s, t) & = A_s (t_s) e^{-\zeta_s (t_s) \omega_s (t_s) \: t} \sin \left (\omega_{d_s }(t_s)  t + \phi_s (t_s) \right ) \\
{\rm where} \,\, 
A_s (t_s) & = \sqrt{\frac {\left ( \dot u_{i_0} + \zeta_s (t_s) \omega_s (t_s) u_{i_0} \right )^2 + \left ( \omega_{d_s }(t_s) u_{i_0} \right)^2} {\omega_{d_s} (t_s)^2}} \\
\text{and} \quad \phi_s (t_s) & = \tan^{-1} {\frac { \omega_{d_s} (t_s) u_{i_0} } {\dot u_{i_0} + \zeta_s (t_s) \omega_s (t_s) u_{i_0}} }
\end{align}
The dynamic response of the system (see \autoref{timed_response}) is obtained from Eq. \eqref{damped_resp}, where a zero initial velocity normalized with $u_{i_0}$ is used. 
The time axis is normalized with the undamped time period of the nominal system 
\begin{equation}
T_0= {\frac {2\pi} {\omega_0} }
\end{equation}
As is seen in \autoref{timed_response}, the difference in the decay rate of the response of the nominal system and the digital twin model is clearly identified. Therefore, by considering the decay rate, it would be possible to estimate $\Delta_m$ from their difference at a certain point in the slow time. This can be done by directly measuring the displacement or velocity readouts (using optical readout for example) and obtaining two successive peaks as shown in \autoref{timed_response}. To that end, we leverage the logarithmic decrement method (see for example, \cite{inman1994engineering}) to obtain the damping factor. The basic premise here is to estimate the damping factor through measurements of the heights of two (or more) successive peaks. For any given damped oscillator, the \emph{logarithmic decrement} is defined as:
\begin{equation}
\delta  = \ln{\frac {u(t)} {u(t+T)} }
\end{equation} 
Using the expression of $u(t)$ as given in Eq. \eqref{damped_resp}, it can be shown that for the digital twin:
\begin{equation} \label{logarithemic decrement}
\delta_m  = {\frac{2\pi \zeta_m}{\sqrt{1-\zeta_m^2}}}.
\end{equation} 
Similarly, the logarithmic decrement for the nominal system $\delta_0$ can be expressed by replacing $\zeta_m$ with the $\zeta_0$ in Eq. \eqref{logarithemic decrement}. Assuming that the logarithmic decrements for the nominal system and the digital twin have been experimentally measured from the response readouts, we propose an approach for obtaining the mass absorption factor ($\Delta_{m}$). We proceed further by taking the ratio of the logarithmic decrements for both the oscillators, which results in the following expression
\begin{equation}
{\frac {\delta_0}  {\delta_m } } = {\frac {\zeta_0} {\zeta_m} } {\frac {\sqrt{1-\zeta_m^2}} {\sqrt{1-\zeta_0^2}} }.
\end{equation} 
Subsequently, taking the square of the above equation yields:
\begin{equation}
\left ({\frac {\delta_0}  {\delta_m } }  \right)^2= {\frac {1+\Delta_m-\zeta_0^2}{1-\zeta_0^2} }. 
\end{equation} 
The mass absorption factor therefore can be obtained by solving this equation as
\begin{equation} \label{ld_mass_id}
\Delta_m = \left ( 1- \zeta_0^2 \right ) \left \{ \left ({\frac {\delta_0}  {\delta_m } }  \right)^2 -1 \right \}
\end{equation} 
Although this sensing technique relies on the measurement of the decrement of the dynamic response, small damping is not a requirement for this to be applicable. For systems with even very high Q-factor, the logarithmic decrements can be calculated by measuring response peaks several cycles apart as:
\begin{equation}
\delta  = {\frac 1 n}\ln{\frac {u(t)} {u(t+nT)} }.
\end{equation}   
Here, $n > 1$, is the number of periods the measured peaks are apart.  When the Q-factor of the nominal system is high ($\zeta_0^2 \approx 0$), Eq. \eqref{ld_mass_id} can be further simplified as:
\begin{equation} \label{ld_mass_id_approx}
\Delta_m \approx \  \left ({\frac {\delta_0}  {\delta_m } }  \right)^2 -1  
\end{equation} 
%====================================================================
So far, we have discussed the numerical setup to capture the variation of mass in the digital twin model of the SDOF dynamical system. However, such an analytical approach has several limitations. Firstly, in a realistic scenario, the data collected are always contaminated by a certain percentage of noise. The analytical framework proposed above only works for clean data. In other words, this only works in a theoretical setting. Secondly, the analytical can update the system parameter; however, it cannot predict the variation of system parameters with time. In other words, a digital twin developed based on only the formulation above will not be predictive. Last but not least, the formulation proposed approach cannot account for data of multiple fidelity. To address these challenges, we propose to combine the MF-HPCFE proposed in the previous section with the analytical formulation presented above. In essence, the formulation presented above is used as a data processing tool for converting the time-history measurement to mass degradation (or parameter degradation to be more specific). Data of fidelity are processed using the same process and hence, we obtain parameter estimates of different fidelity. The MF-HPCFE is then used to track the temporal evolution of the parameters. An algorithm depicting the steps involved is shown in Algorithm \ref{alg:mf-dt}. 

\begin{algorithm}[ht!]
\caption{The proposed enhanced digital twin}\label{alg:mf-dt}
\textbf{Requirements:} Nominal model
\begin{algorithmic}[1]
\State{Collect data from sensors of different fidelity} \label{step1}
\State{Process the low-fidelity and high-fidelity data to obtain the evolution of system parameters (low-fidelity and high-fidelity estimates) with time (See
Section \ref{time_domain} {\slash}  Section \ref{freq_domain})}\label{step2}
\State{Use MF-HPCFE to learn the evolution of the system parameters (see algorithm \ref{alg:hpcfe_train}}
\State{Update the system properties of the nominal model such that $m=m_{0}(1+\Delta_{m})$, $k=k_{0}(1+\Delta_{k})$}
\State{Make the decisions regarding the health of the system, maintenance, and service} \label{step5}
\State{Repeat the step \ref{step1} to \ref{step5}}
\end{algorithmic}
\end{algorithm}

With this setup, the digital twin can handle noisy, sparse, and multi-fidelity data to update itself.  Additionally, MF-HPCFE being a Bayesian surrogate provide an estimate of the predictive uncertainty and entails trust in the digital twin prediction. Overall, the final outcome is a trustworthy, uncertainty aware digital twin that can tackle data of multiple fidelity.

\begin{figure}[ht!]
     \centering
     \includegraphics[width= 0.4\textwidth]{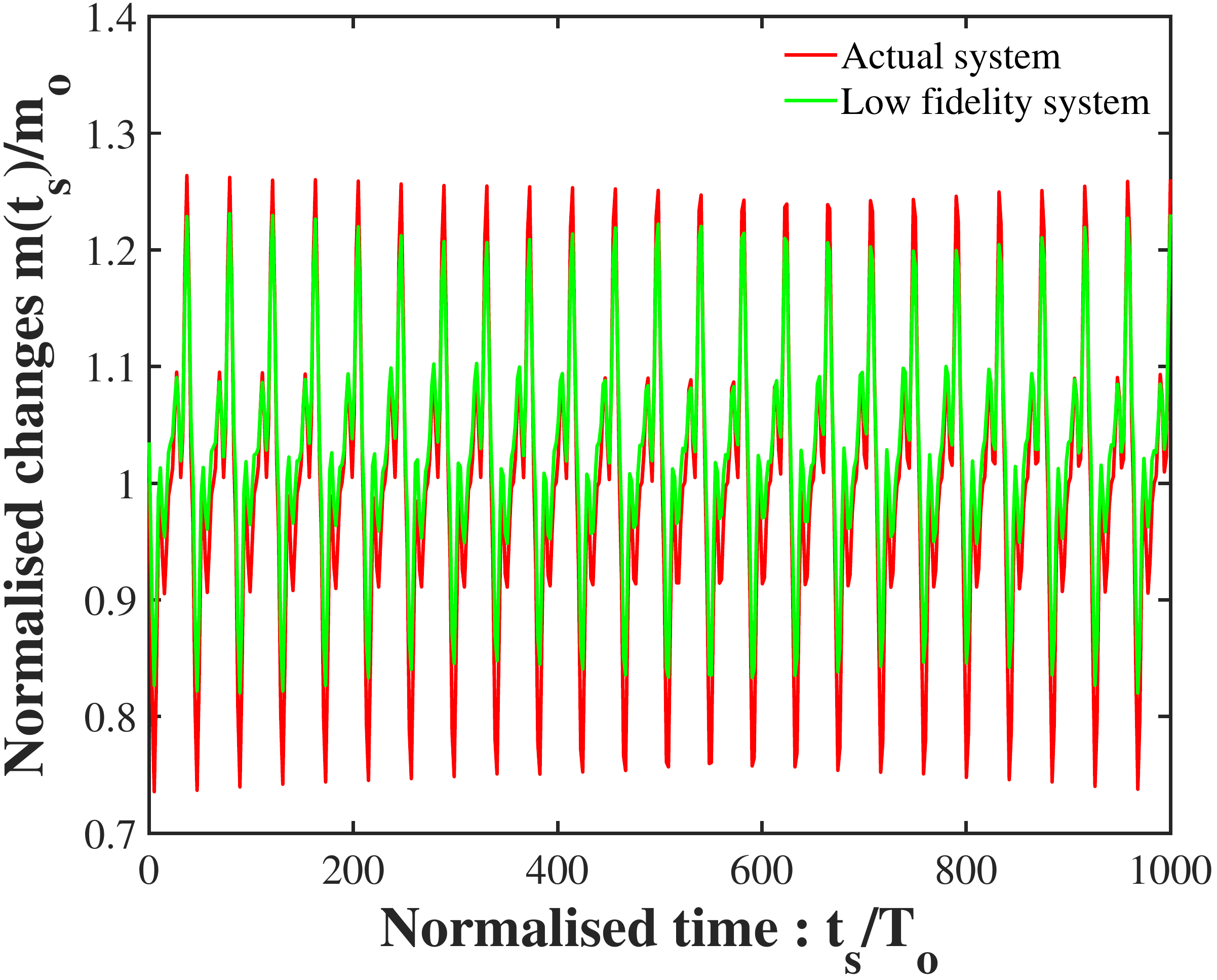}
     \caption{Low fidelity (\autoref{m_fun_GPDT}) and high fidelity (\autoref{var_fnm}) functions representing the mass evolution of the system with normalized time}
     \label{tm_comp}
 \end{figure}

\begin{figure}[ht!]
	\centering
	\subfigure[]
	{\includegraphics[width= 0.45\textwidth]{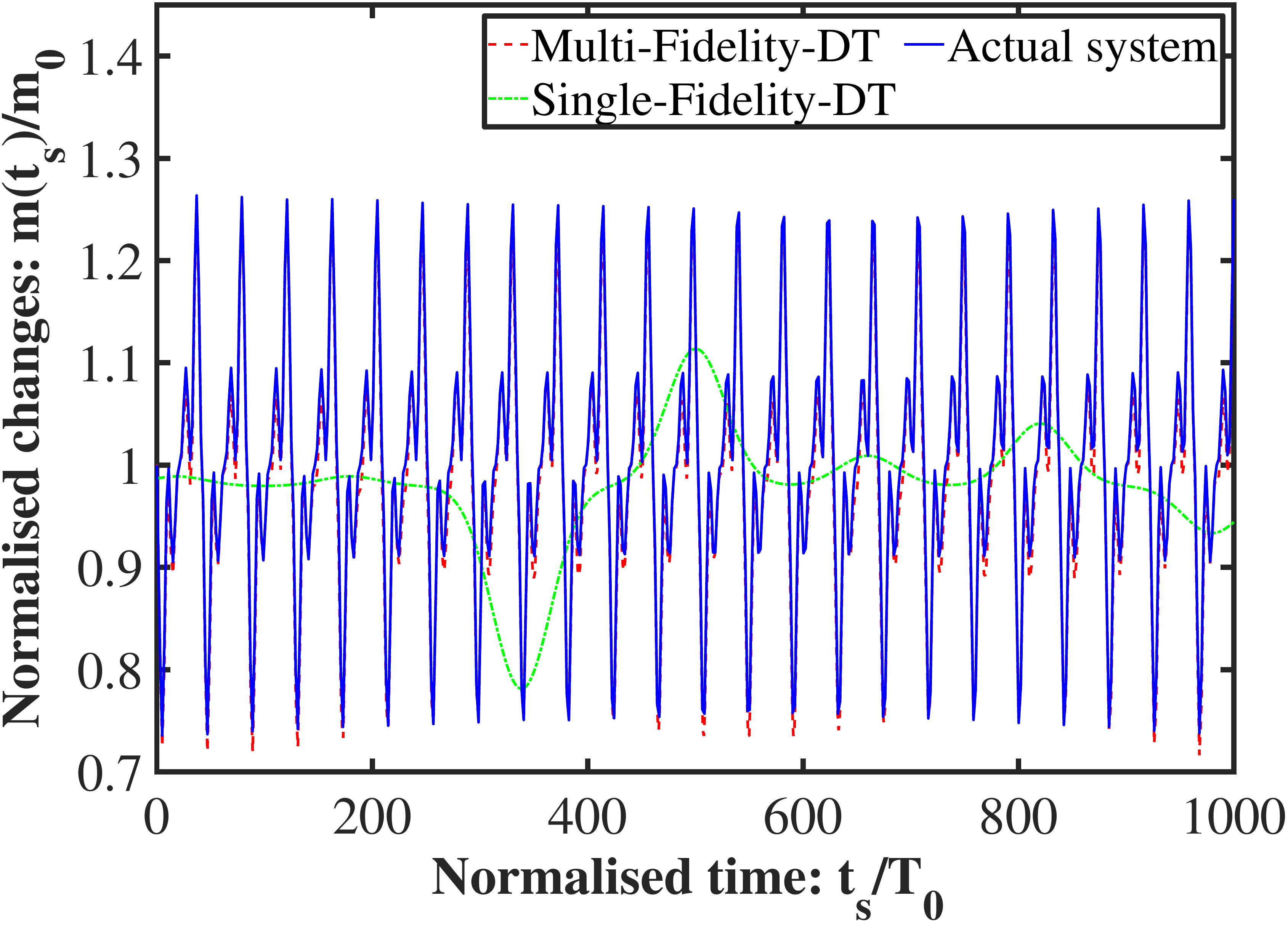}} 
	\subfigure[]
	{\includegraphics[width= 0.45\textwidth]{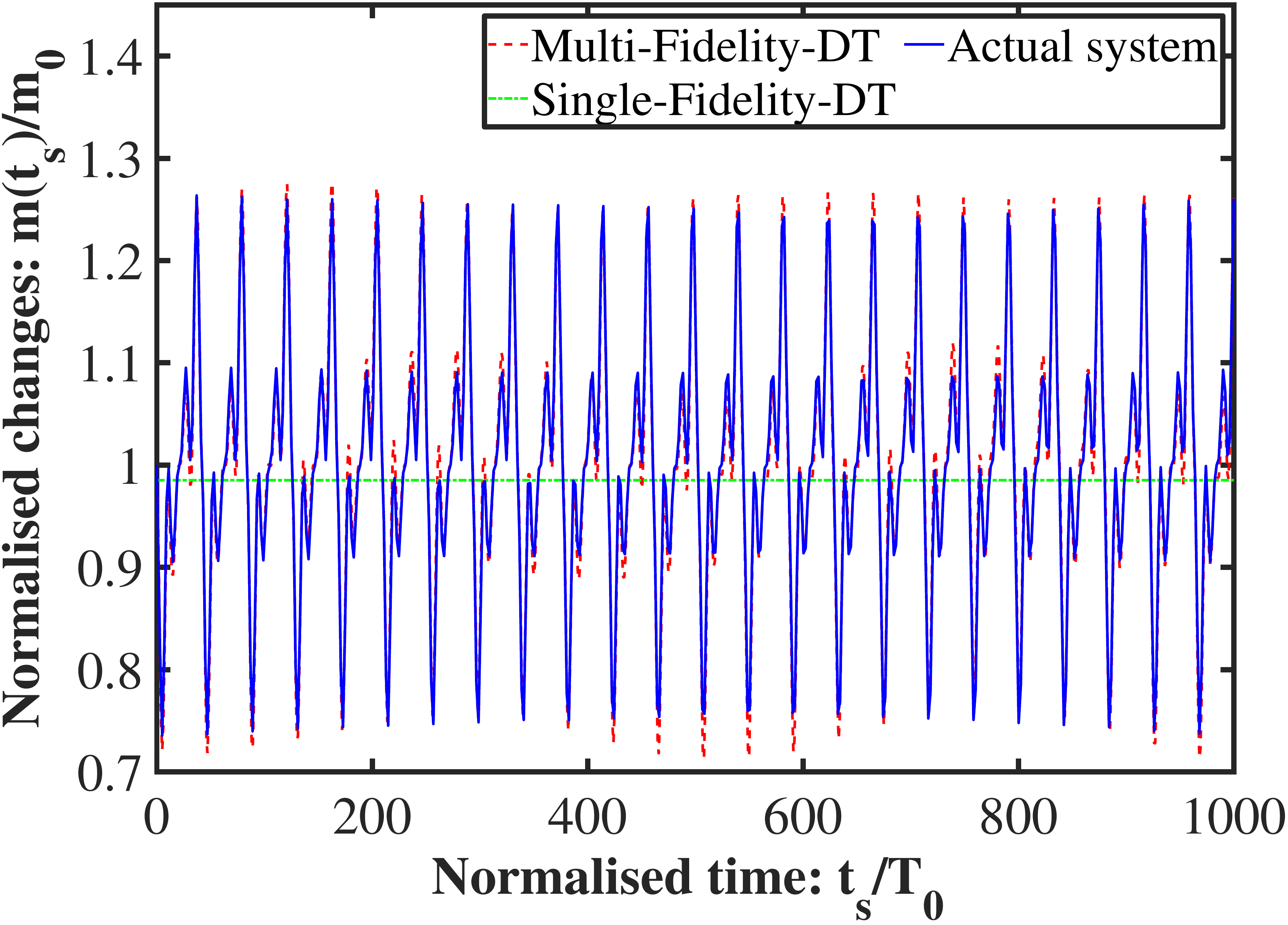}}
         \subfigure[]
	{\includegraphics[width= 0.45\textwidth]{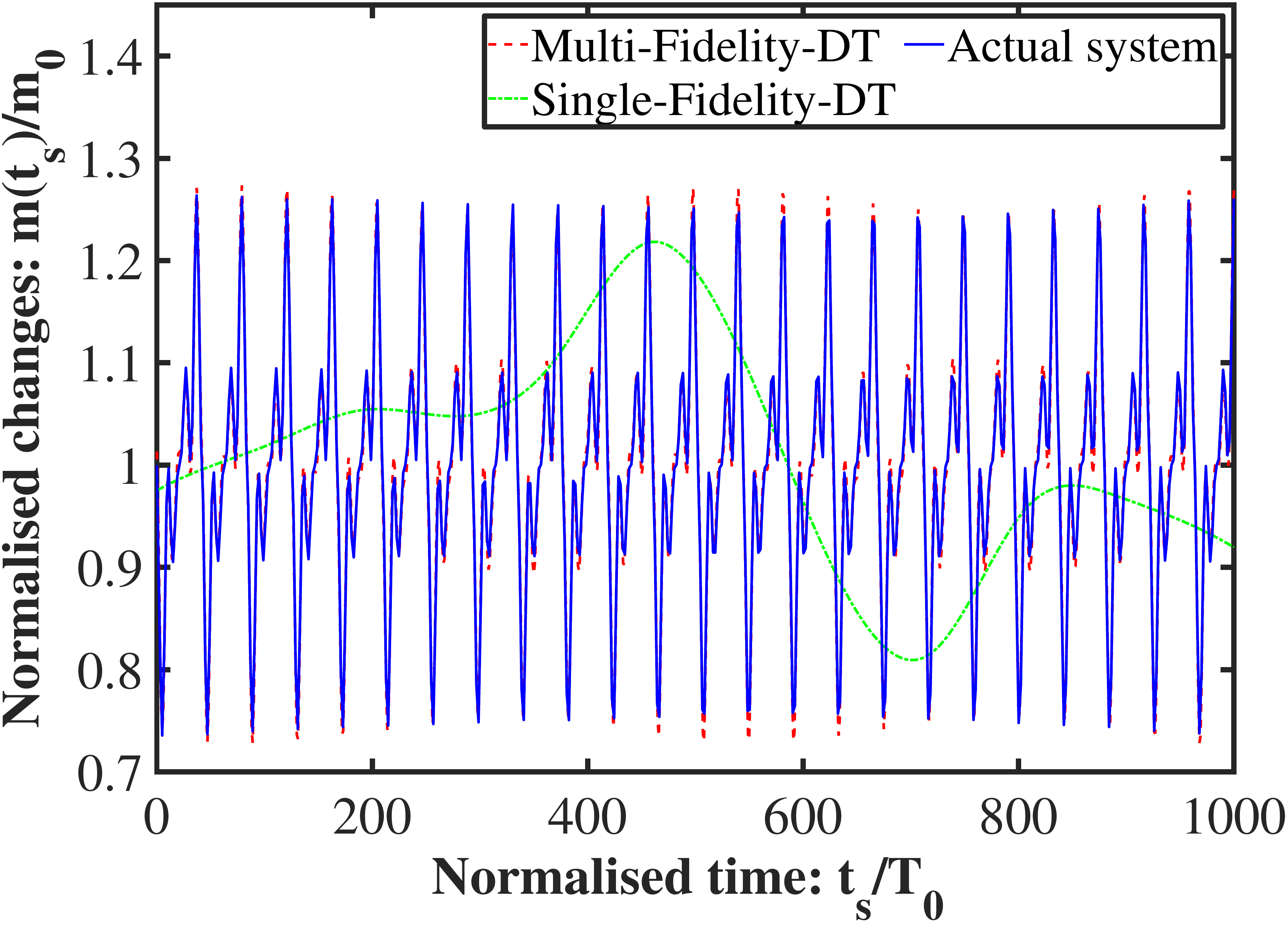}}
        \subfigure[]
	{\includegraphics[width= 0.45\textwidth]{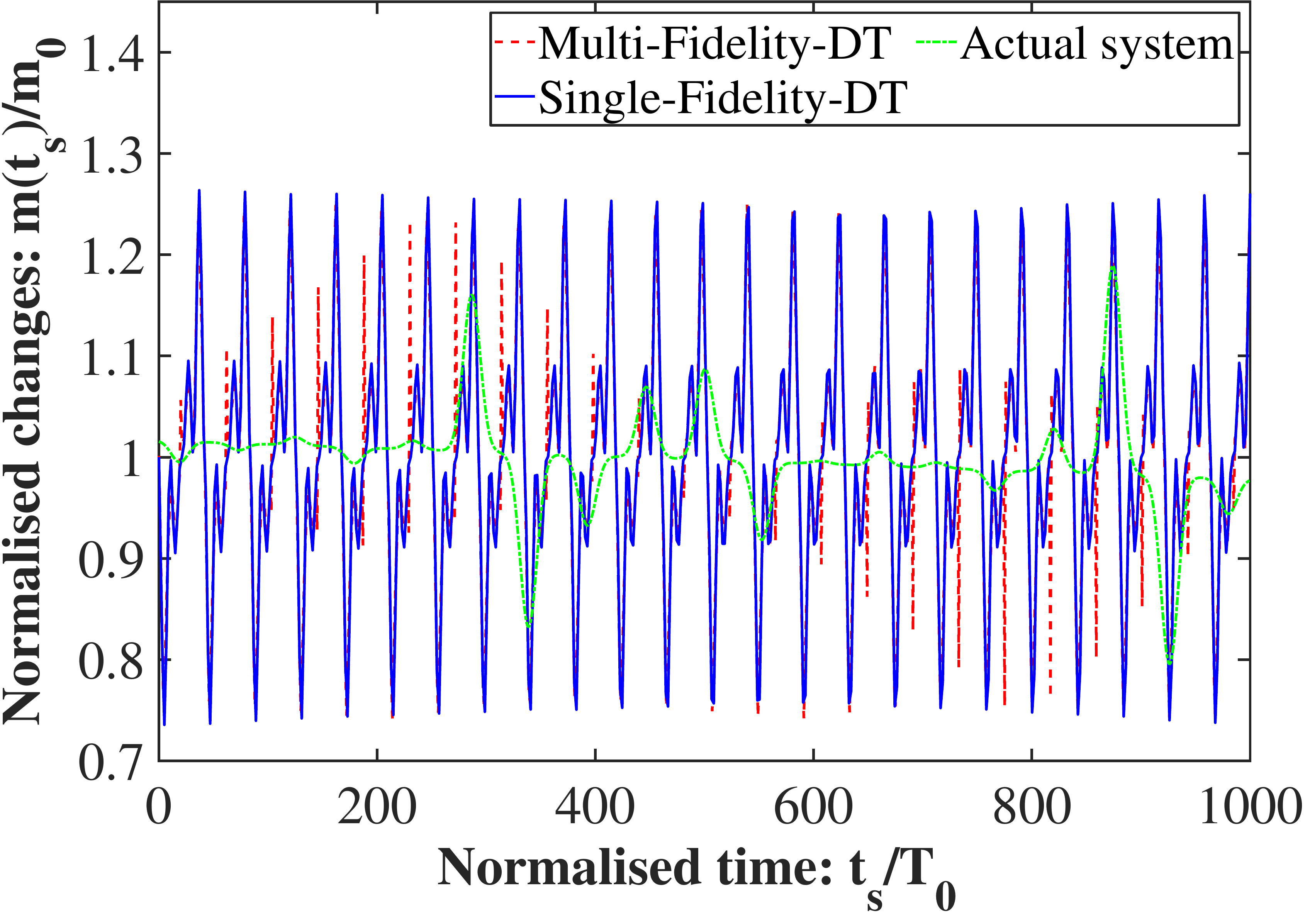}}
	\caption{Prediction results of proposed MF-DT and Single-Fidelity-DT  of mass evolution based on time domain formulations: Normalized response of the oscillator plotted against normalized time $\tau = t/T_0$. The results show the prediction of  MF-DT trained (a) with 7 HF and 501 LF data points, (b) with 11 HF and  501 LF data points, (c) with 12 HF and 501 LF data points, and (d) with 19 HF and 501 LF data points, where LF denotes low-fidelity (\autoref{m_fun}), and $HF_2$ denotes high-fidelity}\label{timed_response_massevol}
\end{figure}

Now, we closely examine the results yielded by the proposed approach. As we discussed earlier, for the numerical experiment, the low-fidelity data and high-fidelity data are synthetically generated using Eqs. \eqref{var_fnm} and \eqref{m_fun_GPDT}, respectively. The visualizations of the ground truth and the low fidelity function of $\Delta_m$ are provided in \autoref{tm_comp}. Further, we investigate the performance of the proposed framework for different training scenarios. The results are summarized in \autoref{timed_response_massevol}. Four cases, shown in \autoref{timed_response_massevol}(a-d) illustrate the significance of the high-fidelity data points in the effective training of the proposed approach.
Here we note that the proposed multi-fidelity approach is conveniently denoted as a multi-fidelity digital twin. Along with the ground truth values of $\Delta_m$, the prediction obtained by the single-fidelity digital twin is also compared. The model is trained with 501 low-fidelity data points, whereas the number of high-fidelity data points is varied. As is seen from the results, the proposed model efficiently captures the variation in the mass $(\Delta_{m})$, whereas the single fidelity DT fails to emulate the ground truth function in all the cases presented. 

% 
% \subsubsection{Using \autoref{m_fun_GPDT} as low fidelity data equation}
% \begin{figure}[h!]
%      \centering
%      \includegraphics[scale=0.5]{time_mass_act_GPDT.pdf}
%      \caption{Mass evolution Low and high fidelity system with time domain }
%      \label{tm_comp_GPDT}
%  \end{figure}

% We evaluated results at 13,15,17,19,21 no of high fidelity data points.Results are as follows,

\begin{figure}[ht!]
     \centering
     \includegraphics[width=0.4\textwidth]{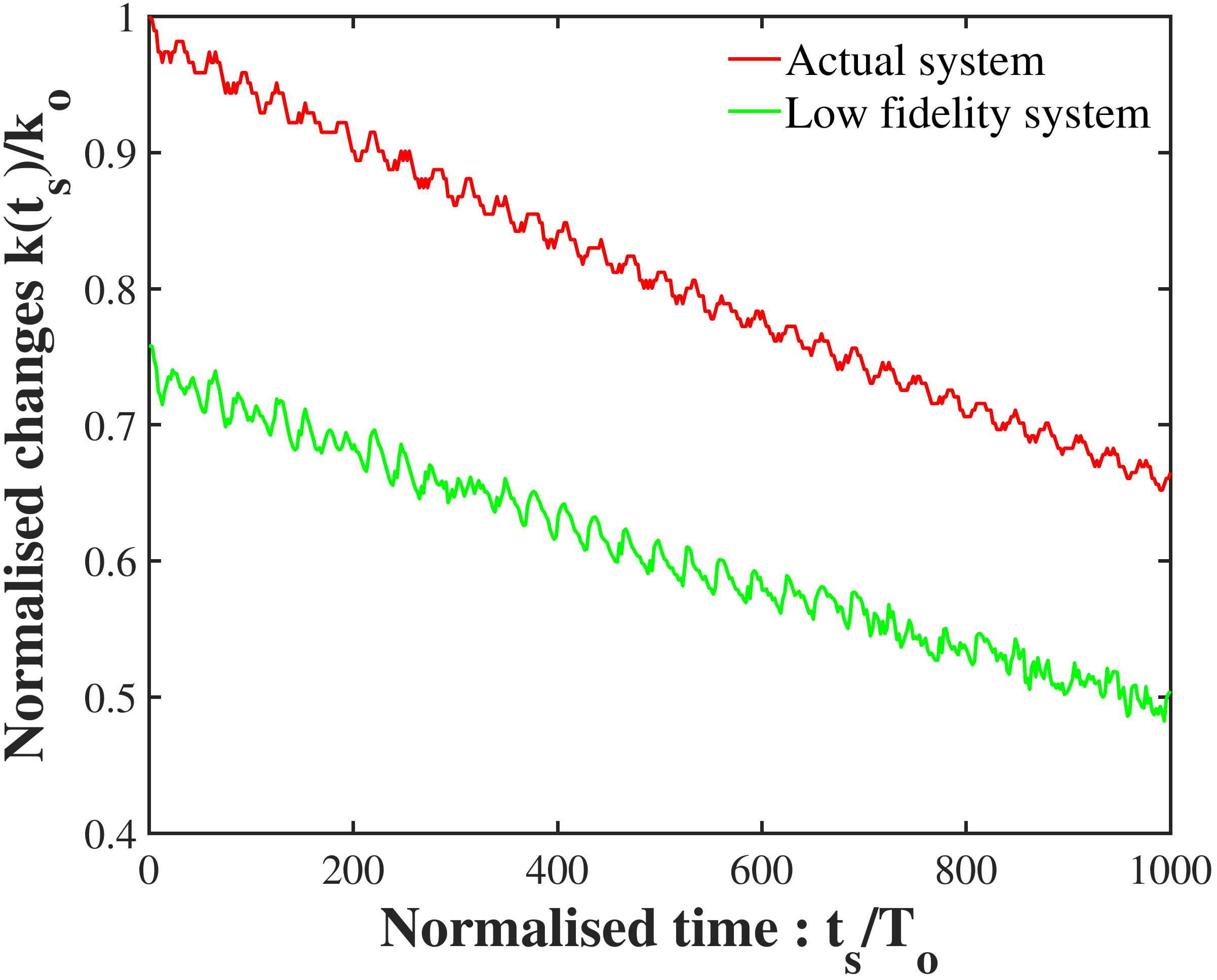}
     \caption{stiffness degradation Low and high fidelity system with time domain }
     \label{tk_comp}
 \end{figure}

\begin{figure}[ht!]
	\centering
	\subfigure[]
	{\includegraphics[width=0.45\textwidth]{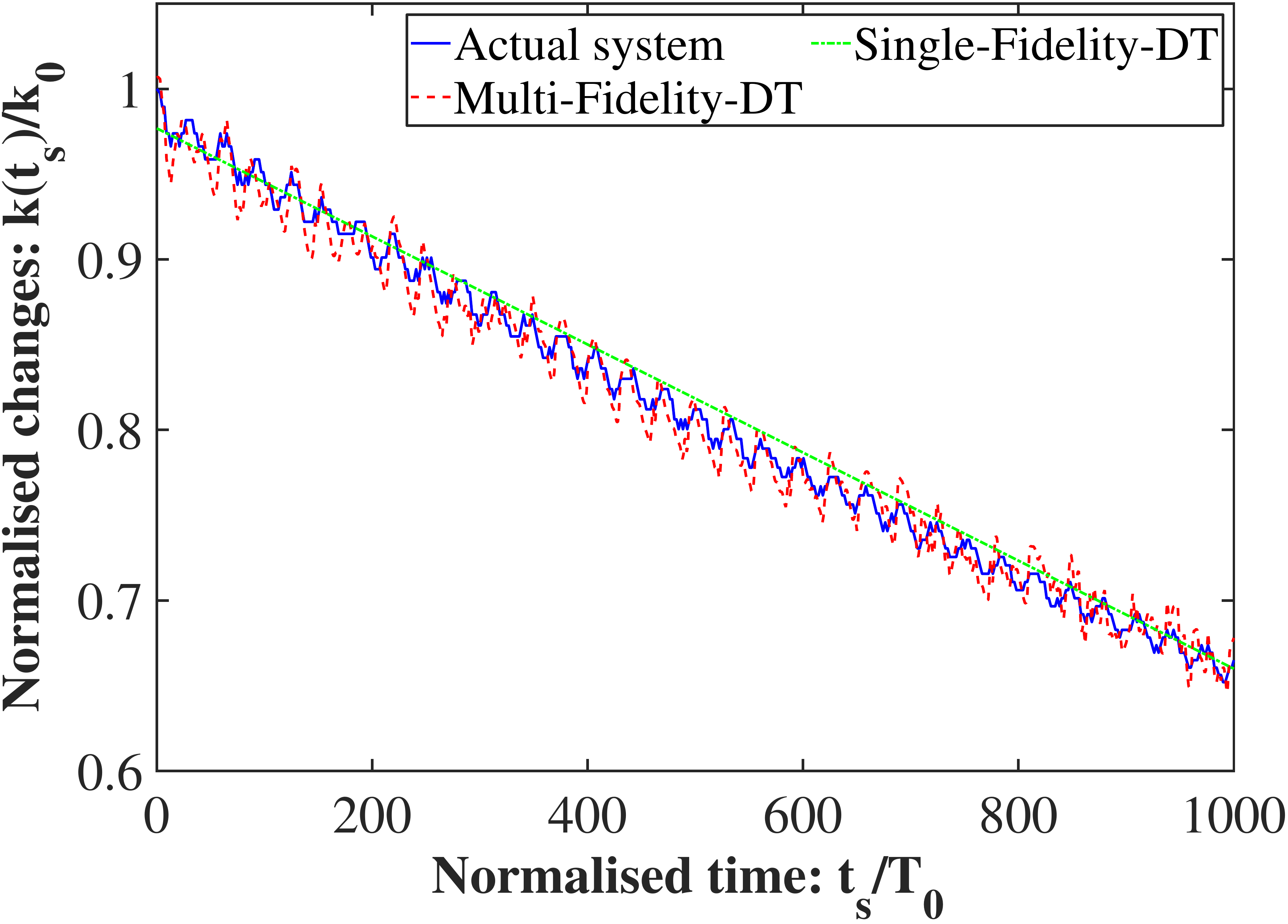}} 
	\subfigure[]
	{\includegraphics[width=0.45\textwidth]{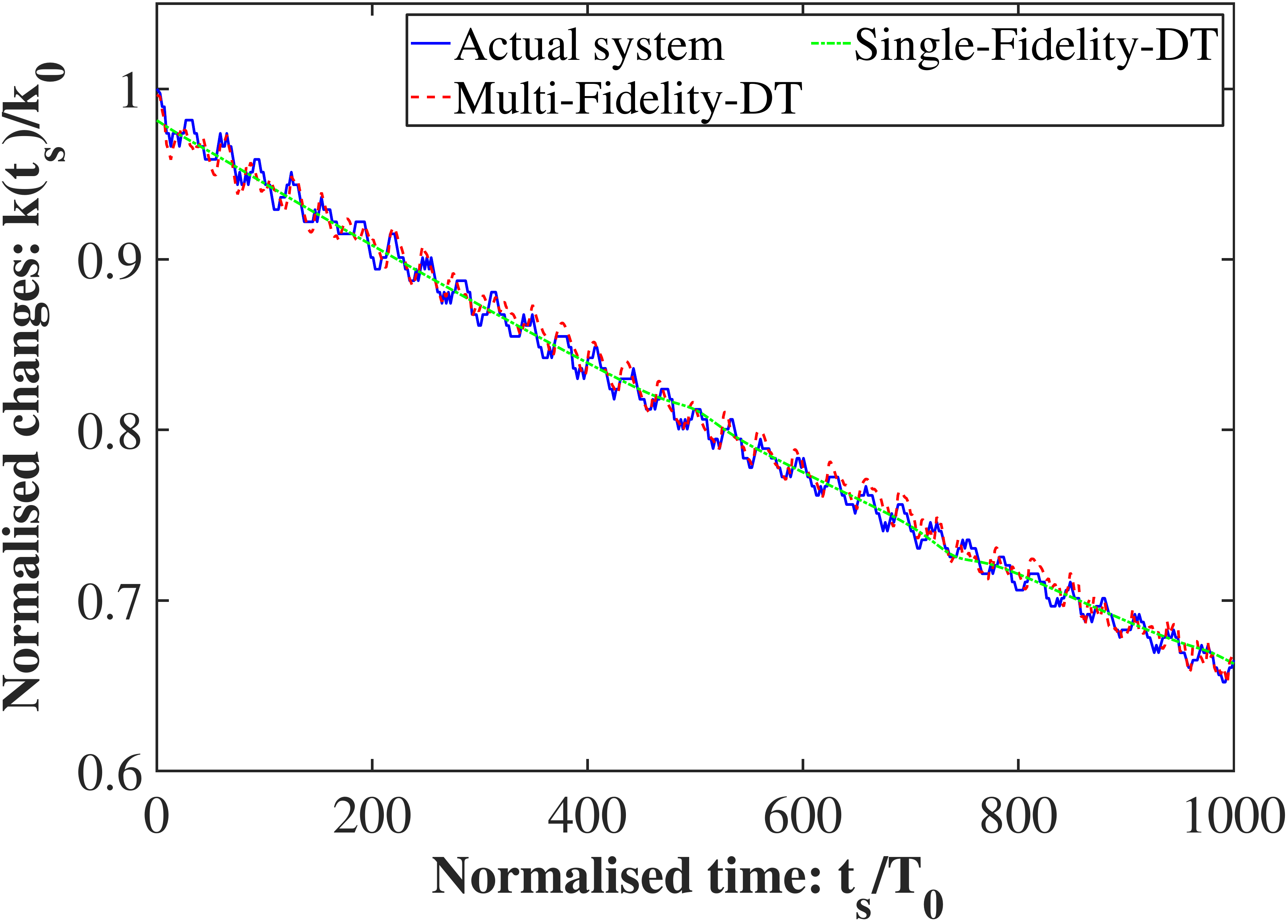}}
         \subfigure[]
	{\includegraphics[width=0.45\textwidth]{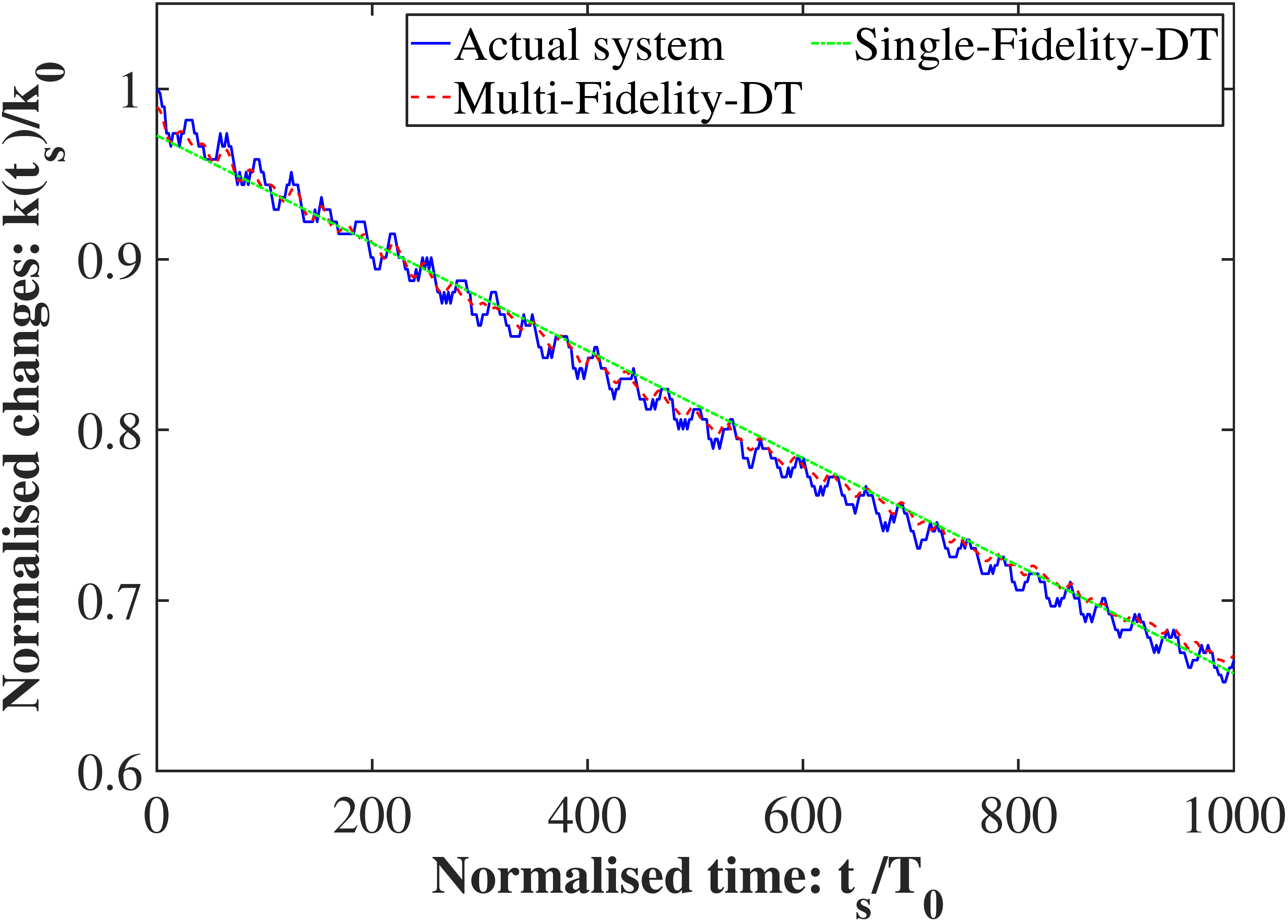}}
        \subfigure[]
	{\includegraphics[width=0.45\textwidth]{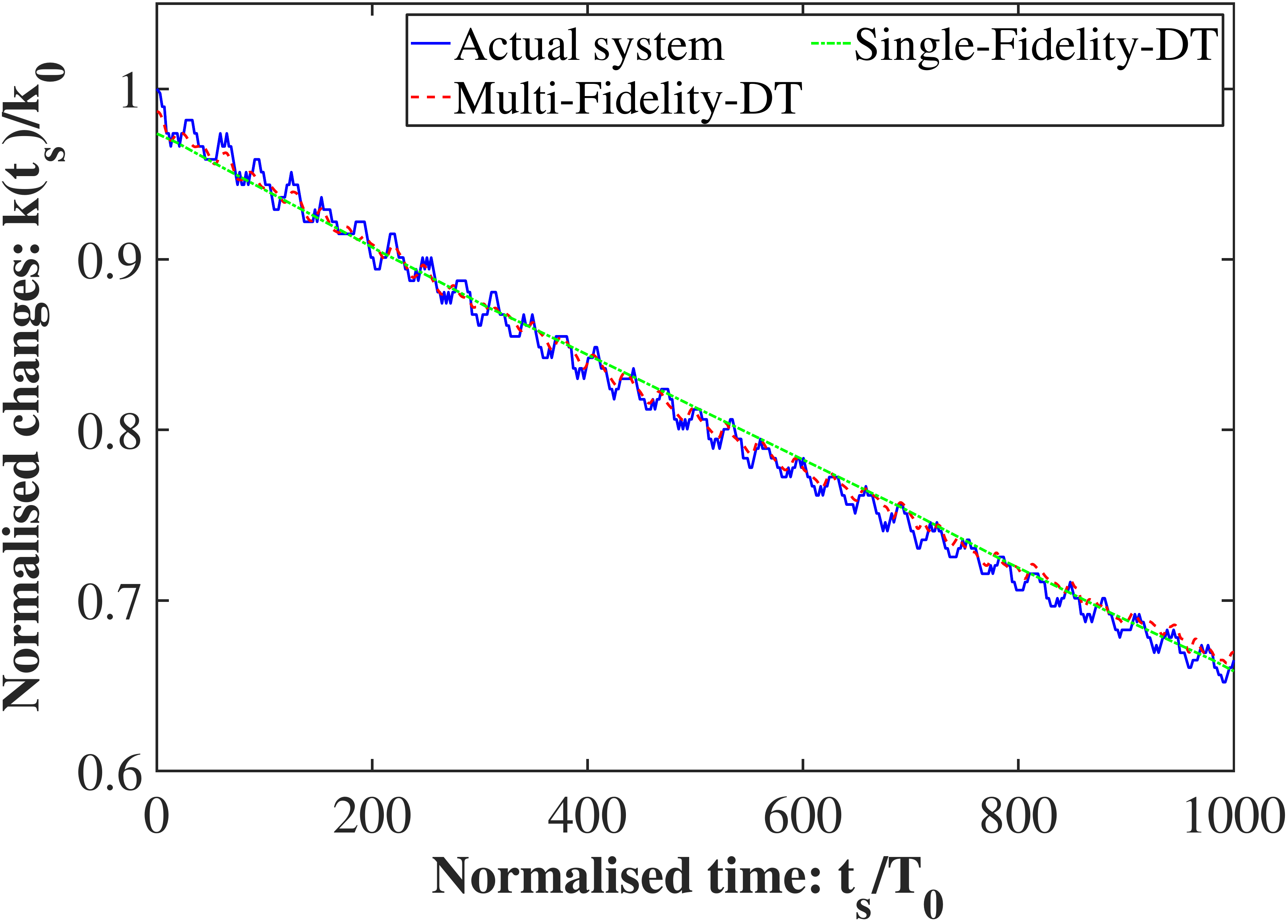}}
	\caption{Prediction results of proposed MF-DT and Single-Fidelity-DT  of stiffness evolution based on time domain formulations: Normalized response of the oscillator plotted against normalized time $\tau = t/T_0$. The results show the prediction of MF-DT trained (a) with 3 HF and  501 LF data points, (b) with 5 HF and  501 LF data points, (c) with 6 HF and 501 LF data points, and (d) with 11 HF and 501 LF data points, where LF denotes low-fidelity (Eq. 38), and HF2 denotes high-fidelity}\label{timed_response_stiffness}
\end{figure}

% \begin{figure}[ht!]
% 	\centering
% 	\subfigure[Figure shows time domain results with 13 HF and  501 LF data points with \autoref{m_fun_GPDT}]
% 	{\includegraphics[width=\subfigwidth]{time_mass_13_GPDT.pdf}} 
% 	\subfigure[Figure shows time domain results with 15 HF and  501 LF data points with \autoref{m_fun_GPDT}]
% 	{\includegraphics[width=\subfigwidth]{time_mass_15_GPDT.pdf}}
%          \subfigure[Figure shows time domain results with 19 HF and 501 LF data points with \autoref{m_fun_GPDT}]
% 	{\includegraphics[width=\subfigwidth]{time_mass_19_GPDT.pdf}}
%         \subfigure[Figure shows time domain results with 21 HF and 501 LF data points with \autoref{m_fun_GPDT}]
% 	{\includegraphics[width=\subfigwidth]{time_mass_21_GPDT.pdf}}
% 	\caption{\label{timed_response_mass} Normalised response of the oscillator in the time domain due to an initial displacement plotted as a function of normalized time $\tau = t/T_0$ and damping factor $\zeta_0=0.02$.}
% \end{figure}
% % 
%========================================================================
Now, we consider the second case where the mass and damping of the nominal model are unchanged, and thus, only the time evolution of stiffness properties affects the digital twin. The overall framework is same and hence we directly proceed with the numerical illustration. 
Eqs. \eqref{var_fnm} and \eqref{m_fun_GPDT} are utilized to generate low-fidelity and high-fidelity data evolving stiffness. The variation of the $\Delta_k$ using the low-fidelity function and high-fidelity functions are illustrated in the \autoref{tk_comp}. A further evaluation of the performance of the proposed multi-fidelity digital twin is carried out by varying the high-fidelity points, and the results of the same are summarized in \autoref{timed_response_stiffness}(a-d). For the presented study, the experimental design is chosen such that the low-fidelity points are fixed to 501 while the high-fidelity points are varied from 3 to 11 intermittently. The results apparently indicate that the proposed framework emulates the ground truth better with an increase in the number of high-fidelity points. Here we note that our objective is to obtain the desired accuracy of the framework with a minimum number of high-fidelity training points as we consider the scarcity of high-fidelity points in real-time scenarios. In regards to the performance of single-fidelity digital twin, it fails to yield the desired result, even with an increased number of high-fidelity points, as was observed previously.
 
%=======================================================================
\subsubsection{Digital twin using frequency domain data}\label{freq_domain}
\begin{figure}[ht!]
	\centering
	\subfigure[$Q_0=5$]
	{\includegraphics[width=0.23\textwidth]{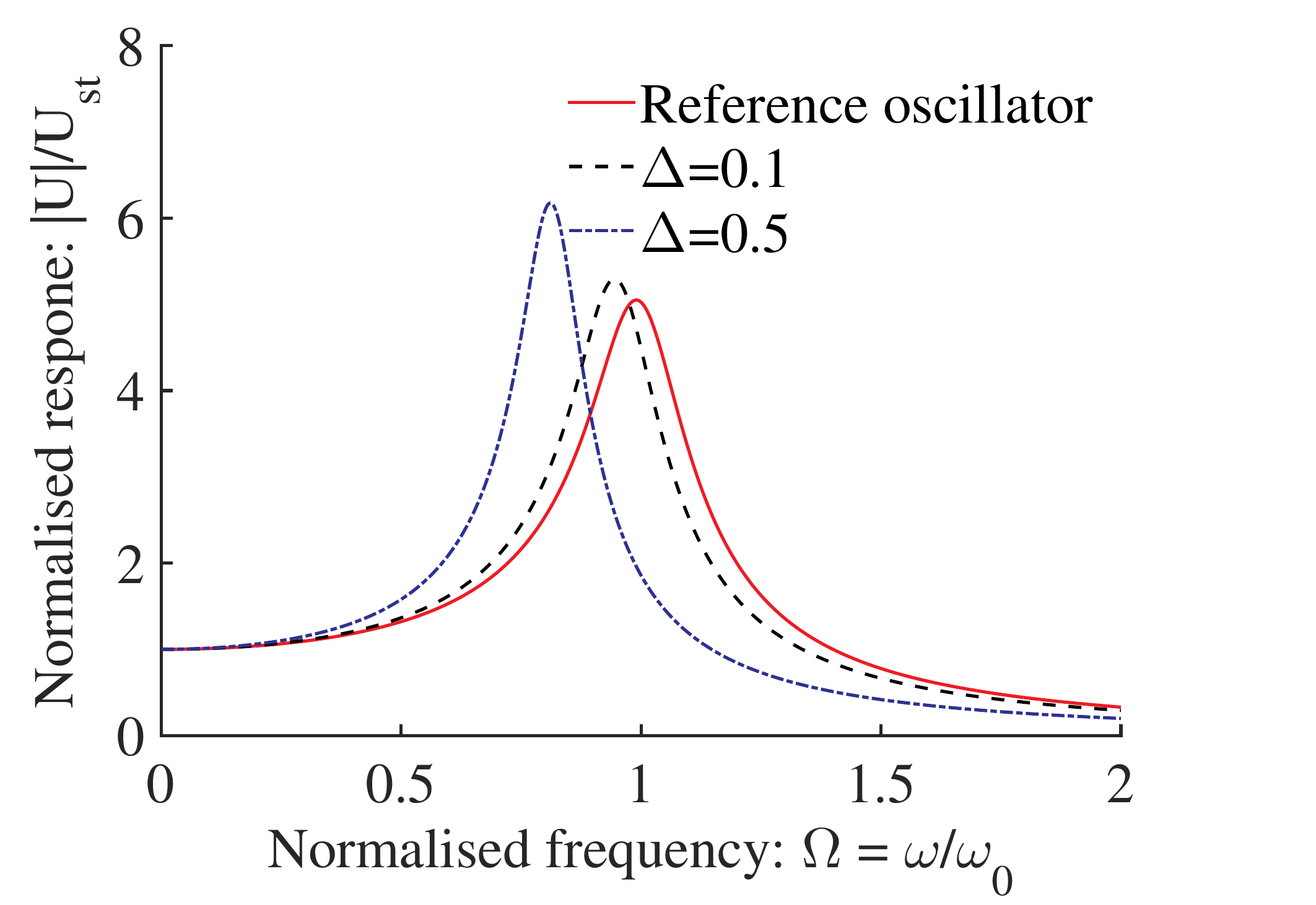}}
	\subfigure[$Q_0=10$]
	{\includegraphics[width=0.23\textwidth]{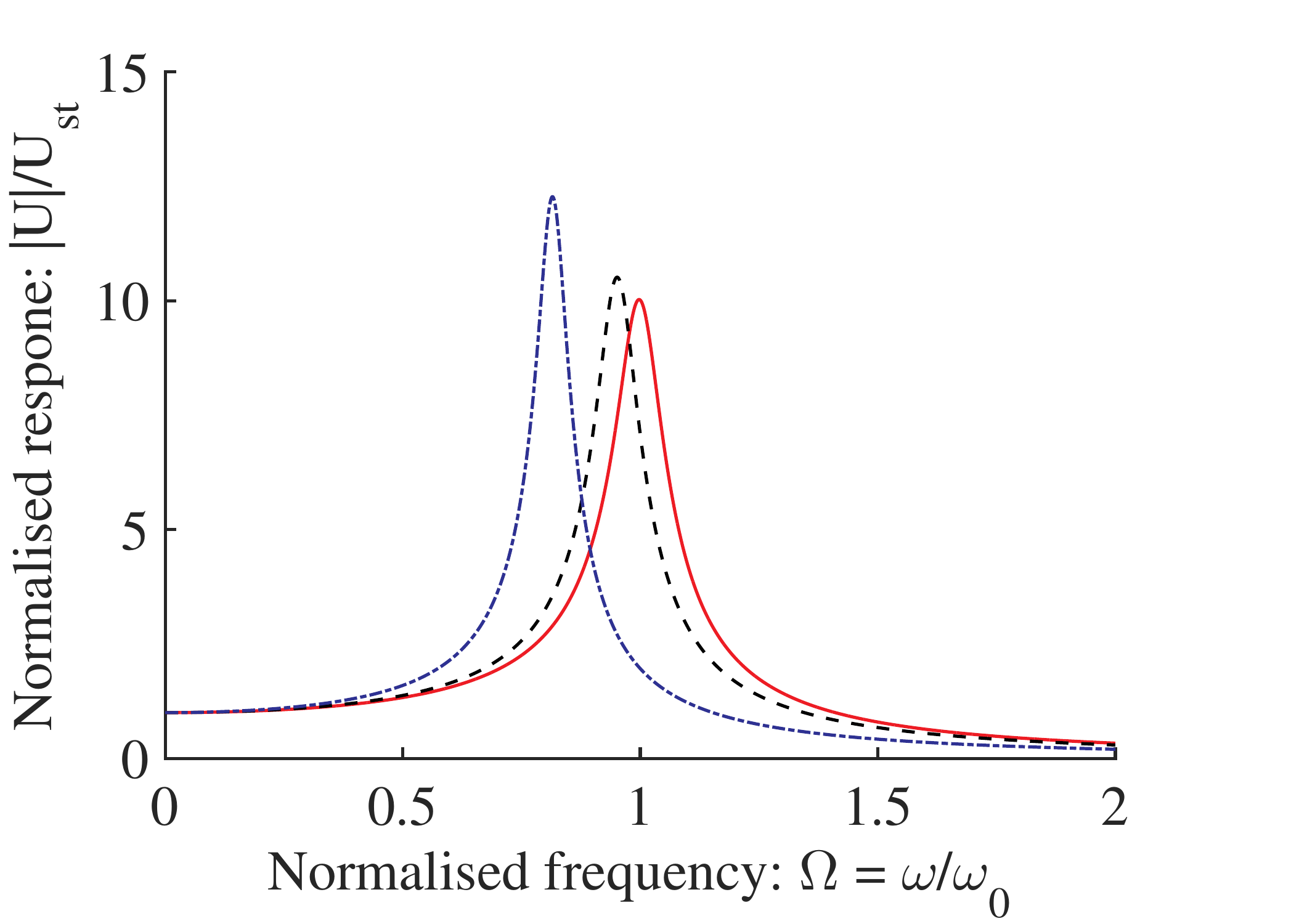}}
	\subfigure[$Q_0=20$]
	{\includegraphics[width=0.23\textwidth]{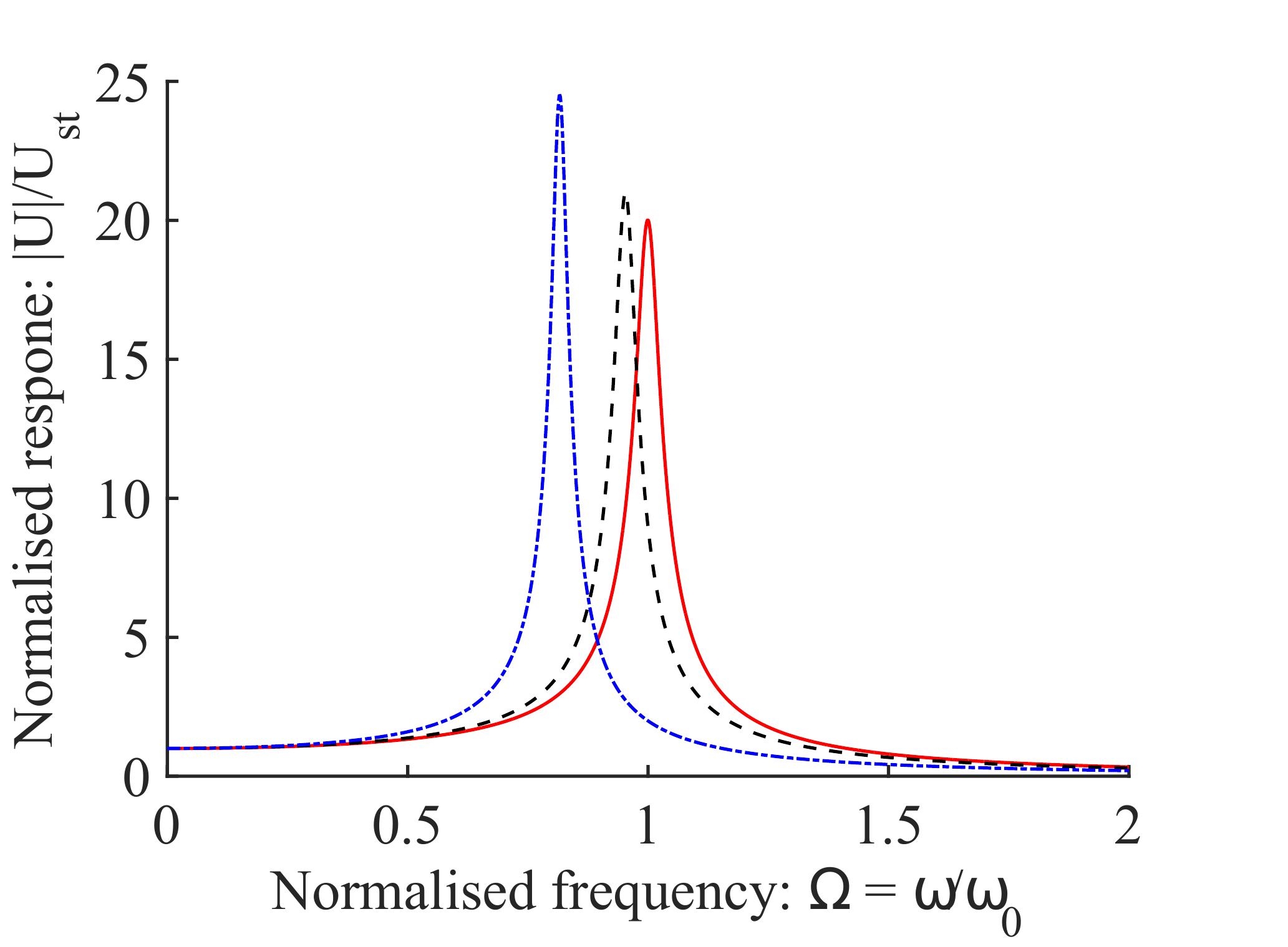}}
	\subfigure[$Q_0=50$]
	{\includegraphics[width=0.23\textwidth]{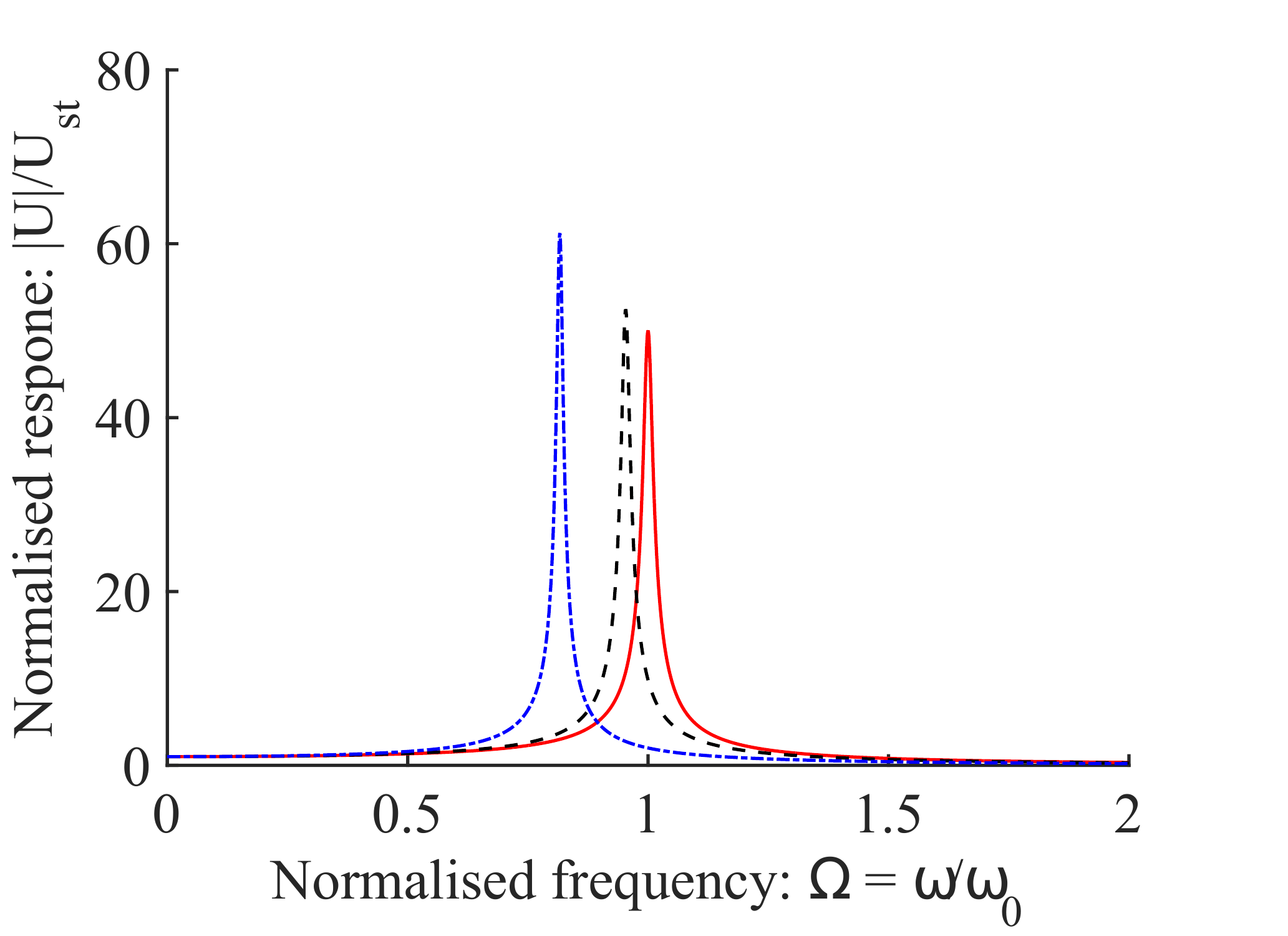}}
	\caption{Normalised response amplitude of the oscillator in the frequency domain as a function of the normalized frequency $\Omega = \omega/\omega_0$ for two different mass absorption factors $\Delta$ and four different Q-factors.}\label{freqd_response}
\end{figure}

When steady-state response due to harmonic or broadband random excitations are considered, the frequency-domain methods provide the most physically intuitive and analytically simplest solutions.  Assuming the amplitude of the harmonic excitation as $F$, from the Laplace transform of \autoref{sdof_eqm}, the response in the frequency domain can be expressed by substituting $s={i}\omega$ as 
\begin{equation}\label{freq_resp1}
\left ( -\omega^2  (1 + \Delta)   + {i} \omega 2 \zeta_0  \omega_0  + \omega_0^2 \right ) U_m({i} \omega) = {\frac { F} {m_0} }
\end{equation}
Dividing this by $\omega_0^2$, the frequency response function of the mass-absorbed oscillator can be expressed as  
\begin{equation}\label{freq_resp2}
U_m({i} \Omega) ={\frac {U_{st} } {-  \Omega^2 (1+\Delta)  + 2 {i} \Omega \zeta_0 + 1} }
\end{equation}
where the normalised frequency and the static response are given by
\begin{equation}
\Omega = {\frac {\omega} {\omega_0} } \quad \text{and} \quad U_{st}={\frac {F} {k} }
\end{equation}
In \autoref{freqd_response}, the frequency response given by \autoref{freq_resp2}  is plotted by normalising it with $U_{st}$. 
It can be observed that the mass-absorbed oscillators show a reduced resonance frequency and reduced damping for all Q-factor values of the reference oscillator. Frequency response functions such as the ones shown in \autoref{freqd_response} can be obtained by the Fast Fourier Transform (FFT) of a measured readout signal in the time domain. In practice, the natural frequency and the damping factor are often obtained from the frequency response function measurements. Therefore, the natural frequency and the damping factor are effectively obtained by `post-processing'  the frequency response functions. In some cases, this process can introduce errors. Here we develop a mass sensing approach that directly uses the frequency response function and avoids direct derivation of the natural frequency and the damping factor.     

\begin{figure}[ht!]
\centering
{\includegraphics[width=0.4\textwidth]{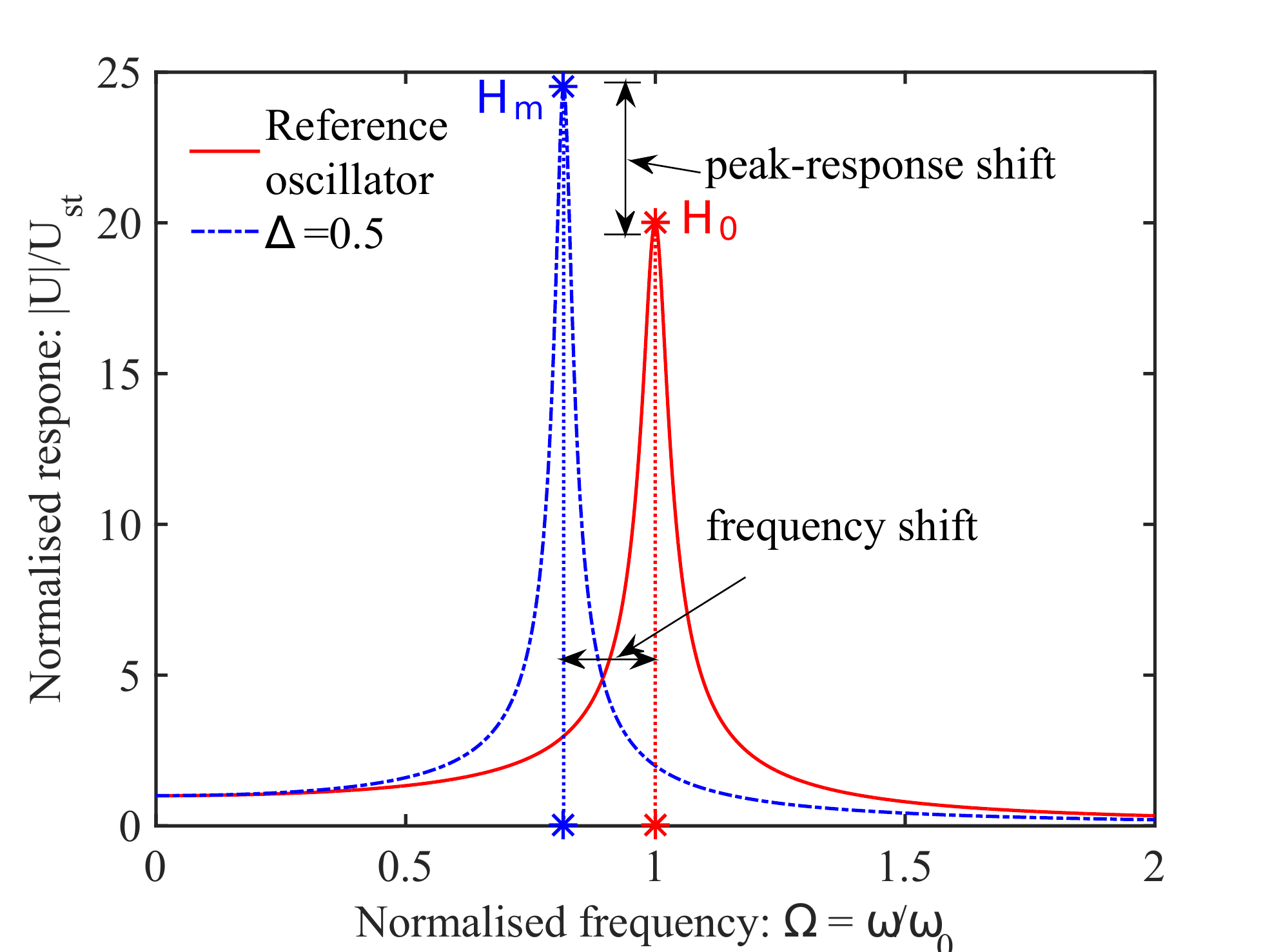}}
\caption{Normalised response amplitude of the oscillator in the frequency domain as a function of the normalised frequency $\Omega = {\omega}/{\omega_0}$ for mass absorption factor $\Delta=0.5$ and $Q_0=20$. $\mathcal{H}_0, \mathcal{H}_m$: Maxima of the frequency response of the reference oscillator and the mass absorbed oscillator.}\label{freqd_response_details}
\end{figure}

We assume that the maxima or the peak of the frequency response function can be located and measured. In \autoref{freqd_response_details}, the peak of the frequency response of the reference oscillator and the mass-absorbed oscillator are shown.
The shift in frequency, as well as damping, are marked in the plot. The aim is to utilize both of this information \emph{simultaneously} to obtain an estimate of the absorbed mass. 

To obtain the maxima of the frequency response, we take a square of the amplitude given by Eq. \eqref{freq_resp2} and set its derivative with respect to $\Omega^2$ to zero, that is
\begin{equation}\label{frf_deri}
\frac{{d}{|U_m|^2}}{{d}{\Omega^2}}=0 
\quad \text{or} \quad
\frac{d}{{\Omega}^2} \left \{ { \frac {1} {  (1-  \Omega^2 (1+\Delta) )^2 + 4 \Omega^2 \zeta_0^2} } \right \}=0
\end{equation}
Solving this equation for $\Omega$, it can be shown that the normalized frequency corresponding to the maximum frequency response amplitude is given by:
\begin{equation}
\Omega_{\max} = {\frac {\sqrt{1+\Delta - 2\zeta_0^2}} {1+\Delta} } = \sqrt{1-2\zeta_m^2}
\end{equation}
The normalized (by $U_{st}$) amplitude of the maximum response at the above frequency point can be obtained as:
\begin{equation}\label{max_hm}
\mathcal{H}_m = | U_m({i} \Omega = {i}\Omega_{\max} ) | / U_{st}= {\frac {1+\Delta} {2\zeta_0\sqrt{1+\Delta-\zeta_0^2} } }
={\frac {1} {2\zeta_m\sqrt{1-\zeta_m^2}} }
\end{equation}
In a similar way the maximum frequency response for the reference oscillator $\mathcal{H}_0$ can also be expressed. Assuming that the maximum frequency response for the reference oscillator and the mass-absorbed oscillator have been experimentally measured, we propose an approach for obtaining the mass absorption factor. Taking the ratio of the maximum frequency response for both of the oscillators we have
\begin{equation}
R_{\mathcal{H}} =  {\frac {\mathcal{H}_m}  {\mathcal{H}_0 } } = (1+\Delta) {\frac {\sqrt{1-\zeta_0^2}} {\sqrt{1+\Delta-\zeta_0^2}} }
\end{equation}         
Squaring both sides, this equation can be simplified to a quadratic equation in  $\Delta$. Solving that equation and keeping only the relevant root,  the mass absorption factor can be obtained as:
\begin{align} \label{Hmax_mass_id}
\Delta & = {\frac {R_{\mathcal{H}}   \left (R_{\mathcal{H}} +\sqrt {R_{\mathcal{H}}^2-4 \zeta_0^2+4 \zeta_0^4} \right)  } 
	{2 \left ( 1- \zeta_0^2 \right ) }
}-1 \\
& = (R_{\mathcal{H}}^2-1) \left (1 + \zeta_0^2 +{\frac {R_{\mathcal{H}}^2+1} {R_{\mathcal{H}}^2} } \zeta_0^4 + \cdots  \right)
\end{align} 
The last equation was obtained by a Taylor series expansion of the expression of $\Delta$ in Eq. \eqref{Hmax_mass_id} about $\zeta_0 = 0$. For systems with high Q-factor, neglecting higher-order terms in $\zeta_0$, the absorbed mass can be explicitly expressed in terms of the peak responses of both the oscillators as
\begin{equation} 
\Delta \approx \left \{ \left ({\frac {\mathcal{H}_m}  {\mathcal{H}_0 } }  \right)^2 -1 \right \} \left (1+ \zeta_0^2 \right )
\end{equation}     
For further lightly damped systems ($\zeta_0^2 \ll 1$) we have the following simplification 
\begin{equation} \label{Hmax_sensing}
\Delta \approx \left ({\frac {\mathcal {H}_m}  {\mathcal {H}_0 } }  \right)^2 -1 \times\alpha
\end{equation}     
Neither the calculation of the resonance frequencies nor the calculation of the damping factors is required to apply this expression. This result is also independent of the Q-factor of the oscillator. Since the frequency response maxima amplitudes appear as a ratio in Eq. \eqref{Hmax_sensing}, units of measurement or normalization do not affect the result. From the point of view of ease of measurement, 
Eq. \eqref{Hmax_sensing} represents the simplest mass sensing approach.
%==========================================================
The above formulations yield the evolution of the digital solely due to the variation in the mass while a similar formulation can be derived for the digital twin model in which only the stiffness property is varied. However, the limitations of the time-domain approach are present here as well; this is addressed by employing the MF-HPCFE as before.

\begin{figure}[ht!]
	\centering
	\subfigure[]
	{\includegraphics[width=0.45\textwidth]{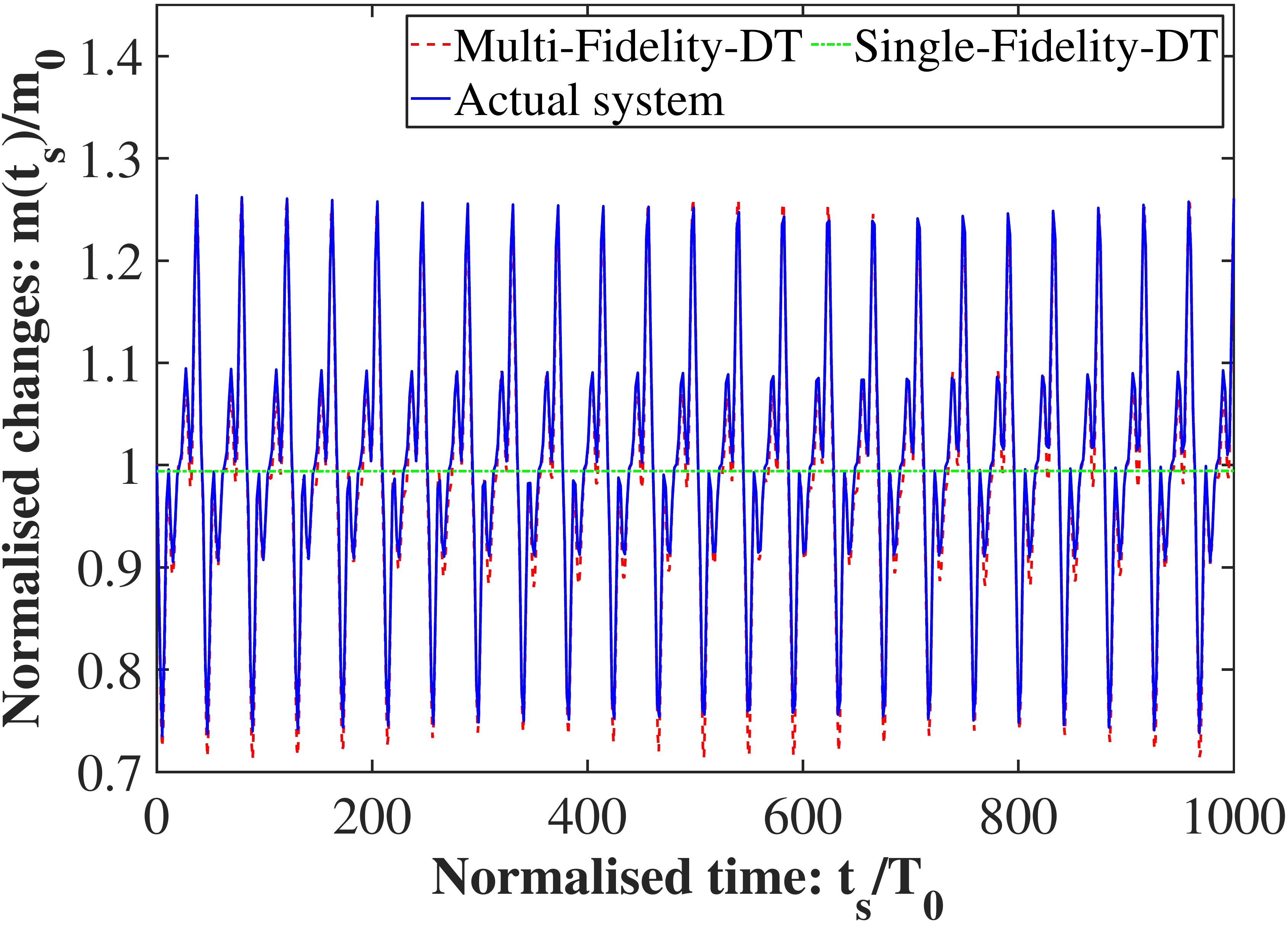}}
	\subfigure[]
	{\includegraphics[width=0.45\textwidth]{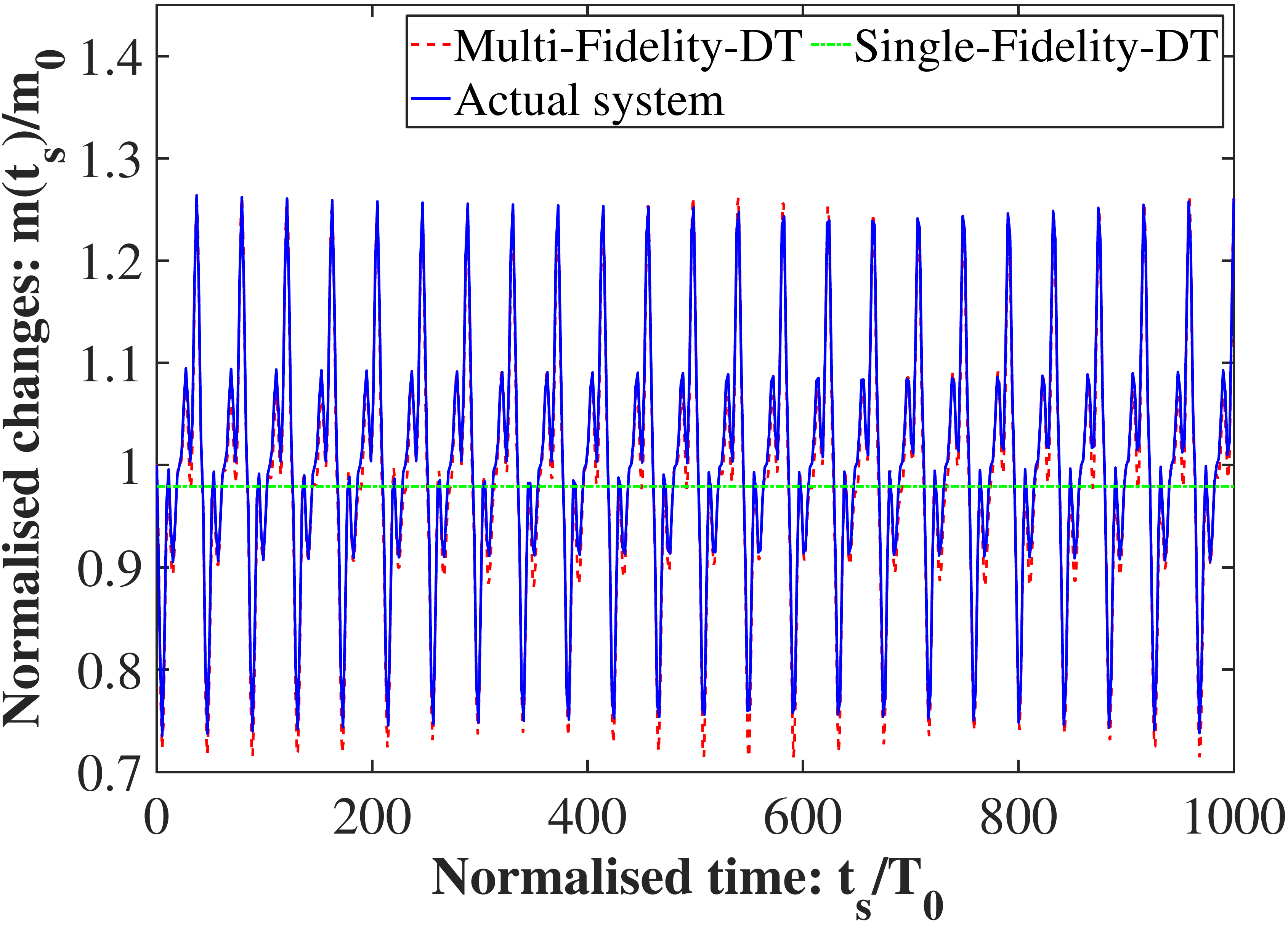}}
	\subfigure[]
	{\includegraphics[width=0.45\textwidth]{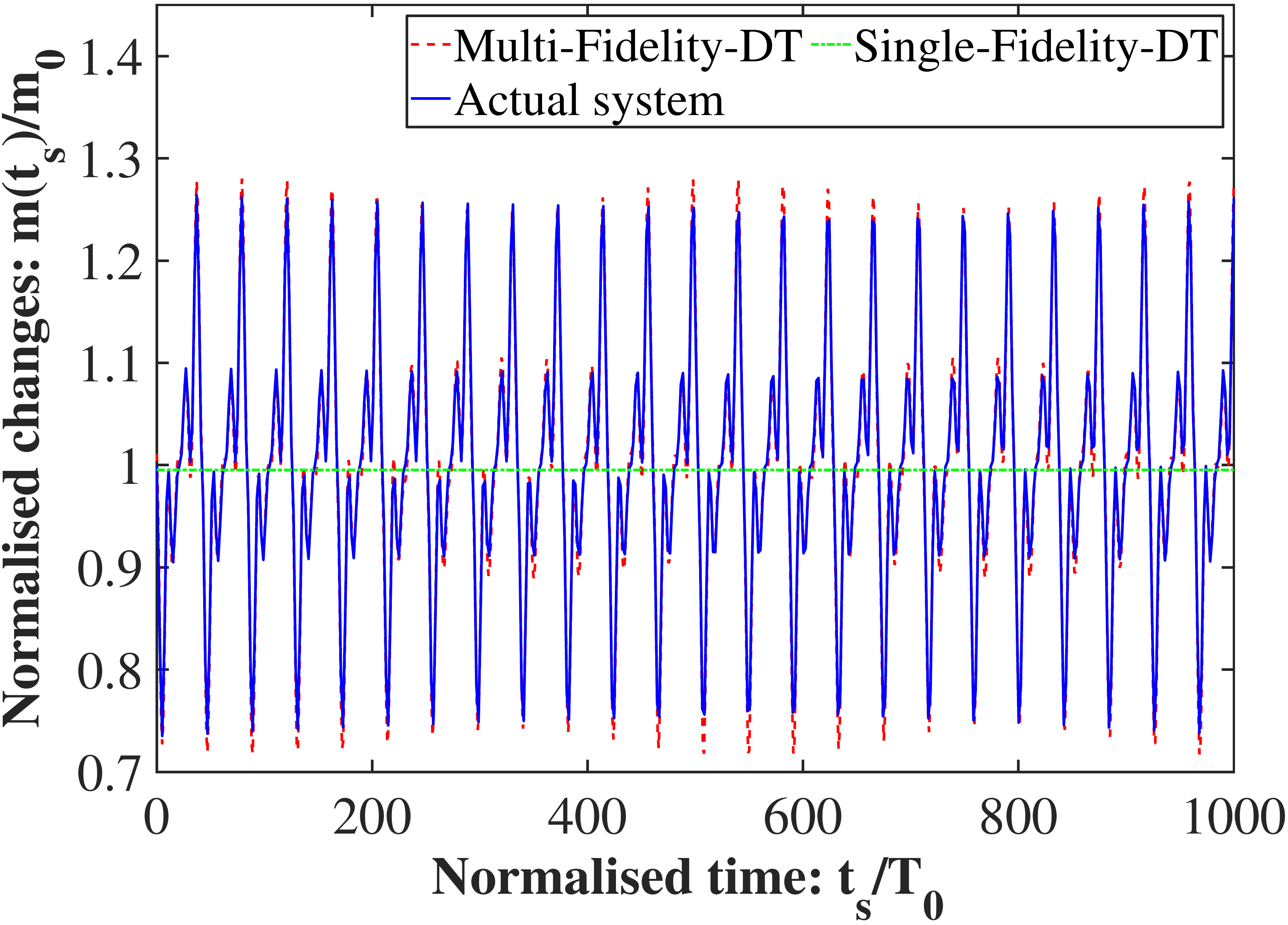}}
	\subfigure[]
	{\includegraphics[width=0.45\textwidth]{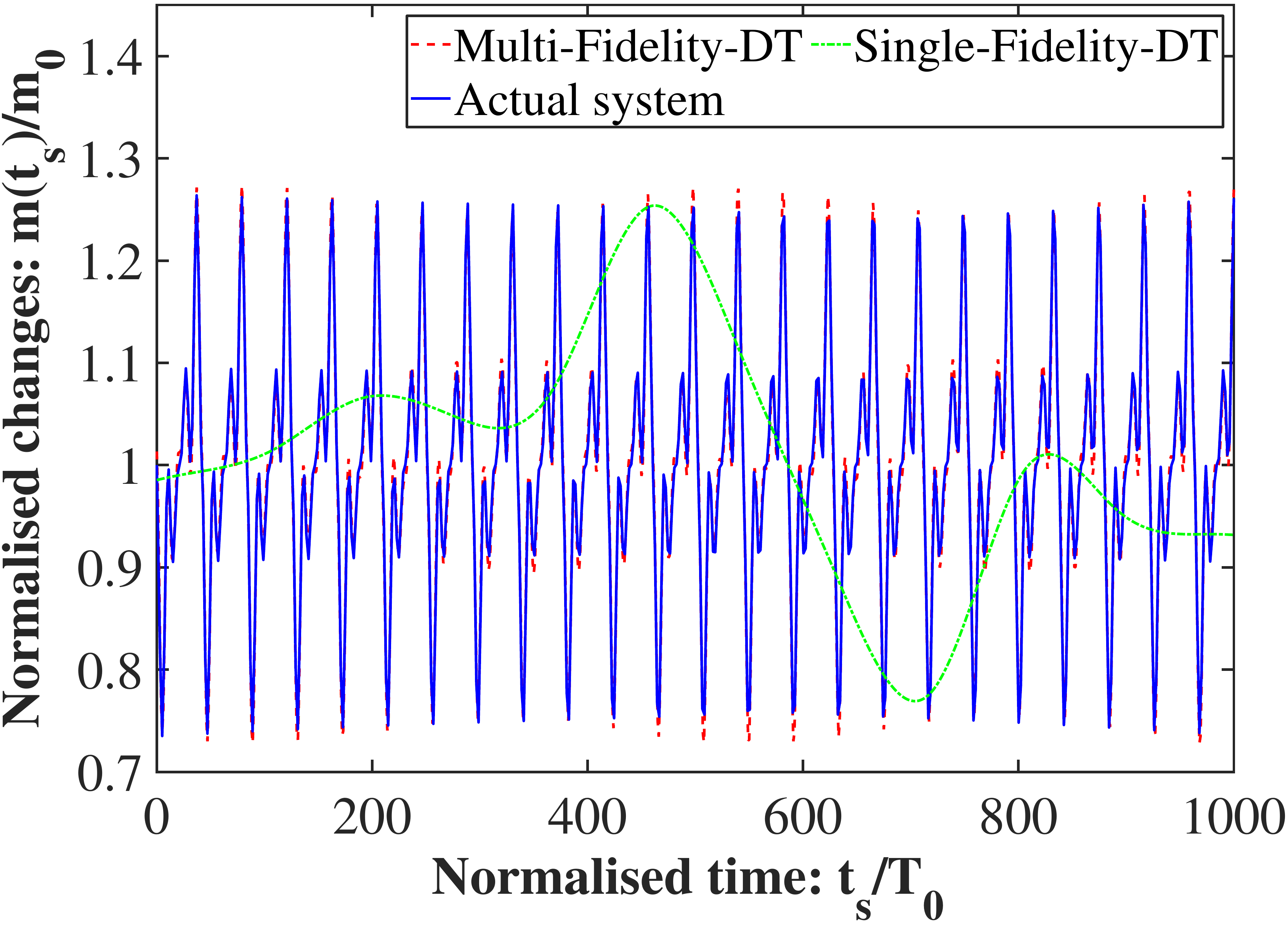}}

\caption{Prediction results of proposed MF-DT and Single-Fidelity-DT  of mass evolution based on frequency domain formulations: Normalized response of the oscillator plotted against normalized time $\tau = t/T_0$. The results show the prediction of  MF-DT trained (a) with 5 HF and 501 LF data points, (b) with  7 HF and 501 LF data points, (c) with 10 HF and 501 LF data points, and (d) with 12 HF and 501 LF data points, where LF denotes low-fidelity (\autoref{m_fun}), and $HF_2$ denotes high-fidelity}\label{freqd_predm}
\end{figure}

Similar to the previous case, i.e., \autoref{time_domain} for the numerical experiment, low-fidelity data and high-fidelity data are synthetically generated utilizing the Eq. \eqref{var_fnm} and \eqref{m_fun_GPDT}.  The results of four different scenarios are showcased in \autoref{freqd_predm}(a-d). The four scenarios are composed such that low-fidelity data points are kept at 501 data points, whereas the number of high-fidelity points is chosen to be 3,7,10, and 12, respectively. It can be apparently observed from the results that the proposed multi-fidelity DT effectively captures the change in the mass $(\Delta m)$. However, the single-fidelity digital twin fails to yield accurate results.
% \begin{figure}[h!]
%      \centering
%      \includegraphics[width=0.8\textwidth]{freq_mass_act.pdf}
%      \caption{Mass evolution Low and high fidelity system with frequency domain }
%      \label{fm_comp}
%  \end{figure}
In digital twin, Mass is evaluated while keeping frequency as input. Similar to the time mass domain, we applied the MF-DT approach to predict the mass degradation function and compared it with the assumed one. We took 501 low-fidelity points from the equation and 3,5,7,9,10,12 high-fidelity data points.
%=====================================================================

\begin{figure}[ht!]
	\centering
	\subfigure[]
	{\includegraphics[width=0.45\textwidth]{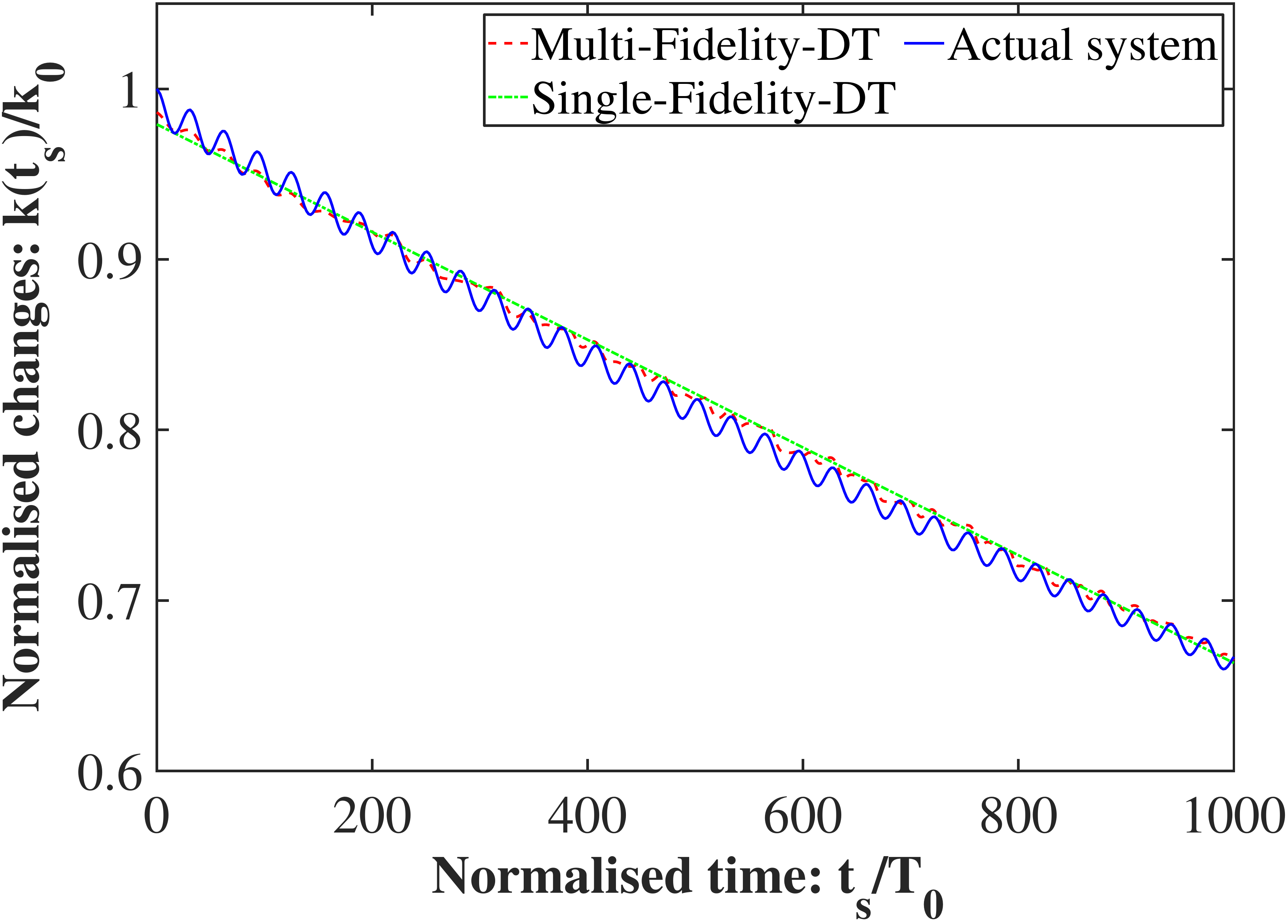}}
	\subfigure[]
	{\includegraphics[width=0.45\textwidth]{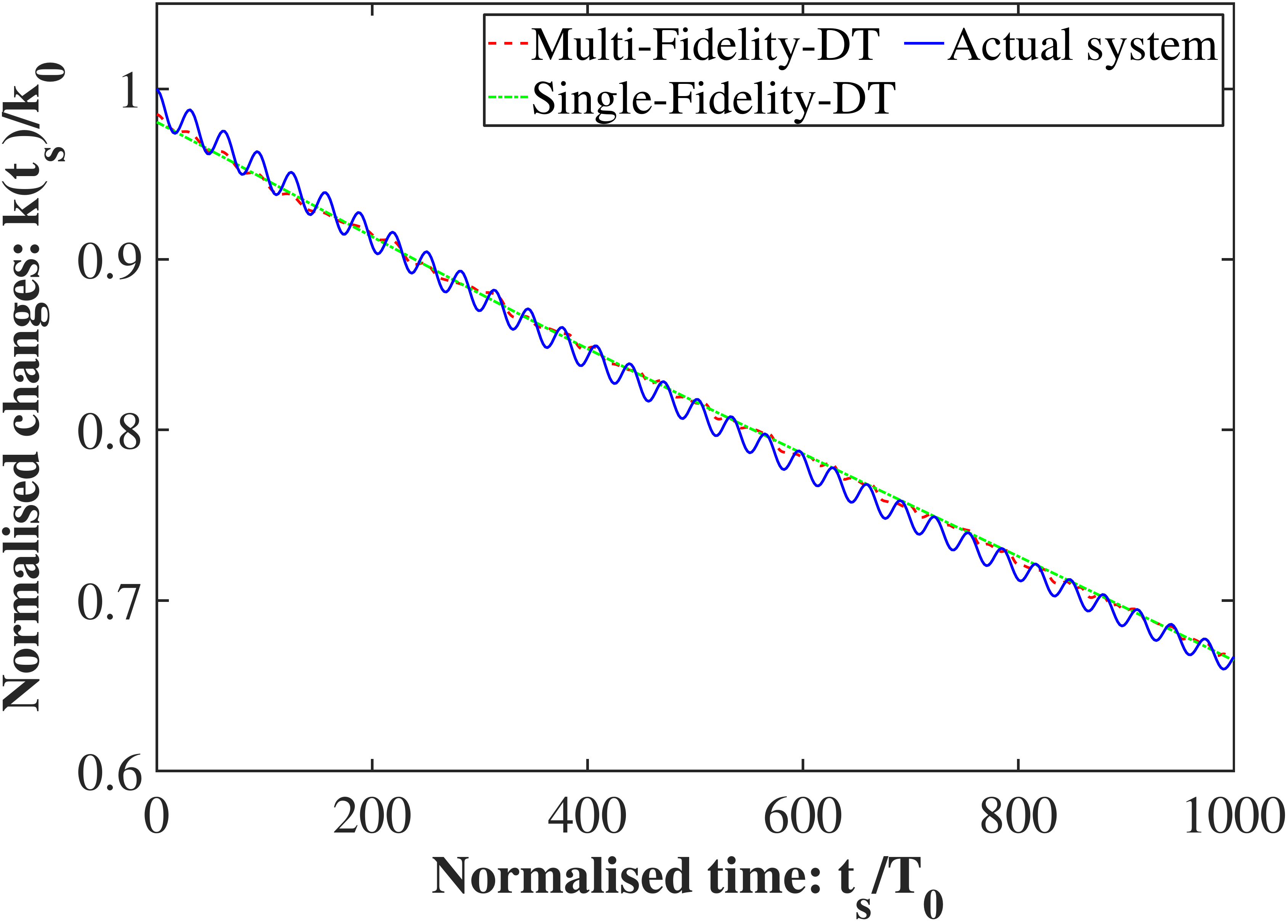}}
	\subfigure[]
	{\includegraphics[width=0.45\textwidth]{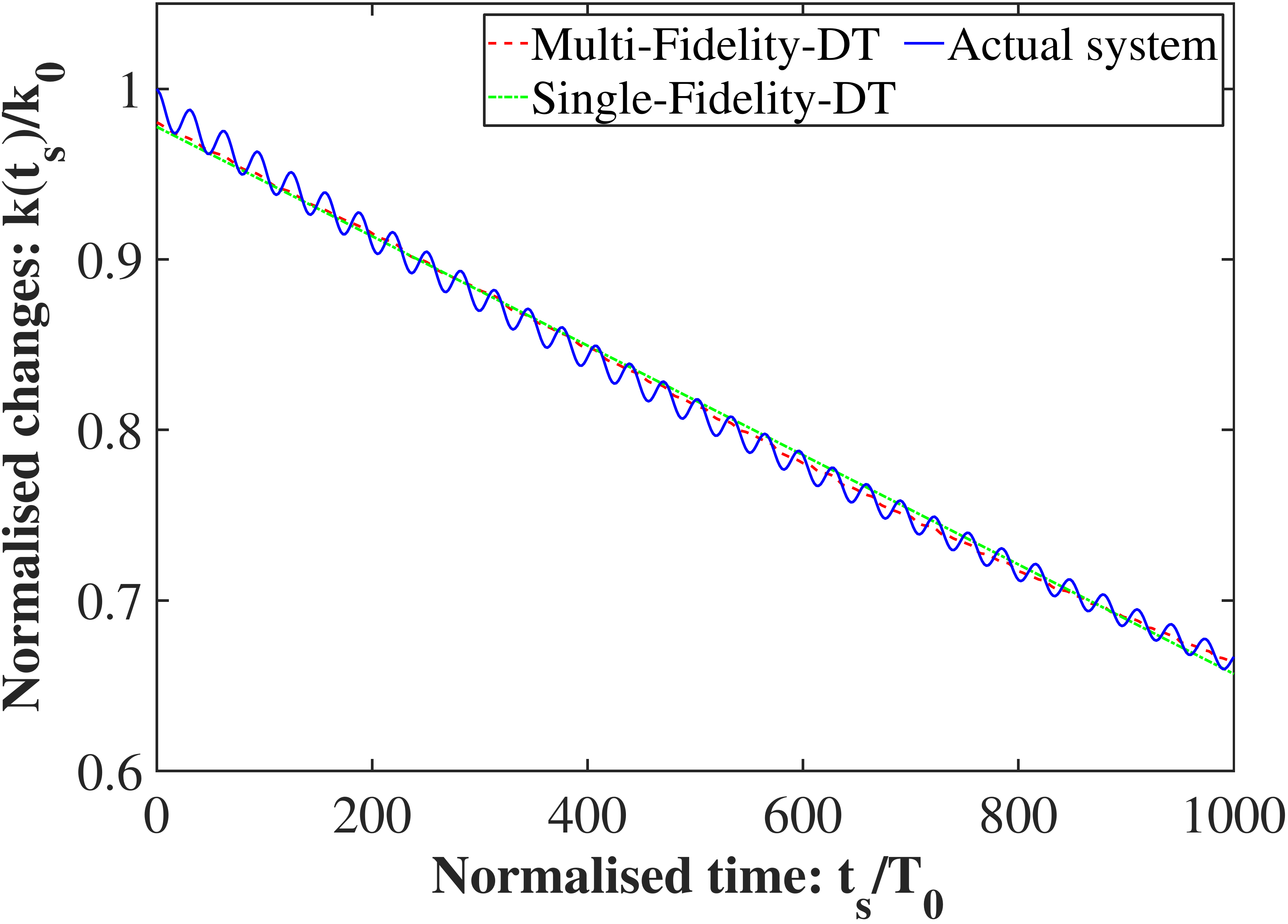}}
	\subfigure[]
	{\includegraphics[width=0.45\textwidth]{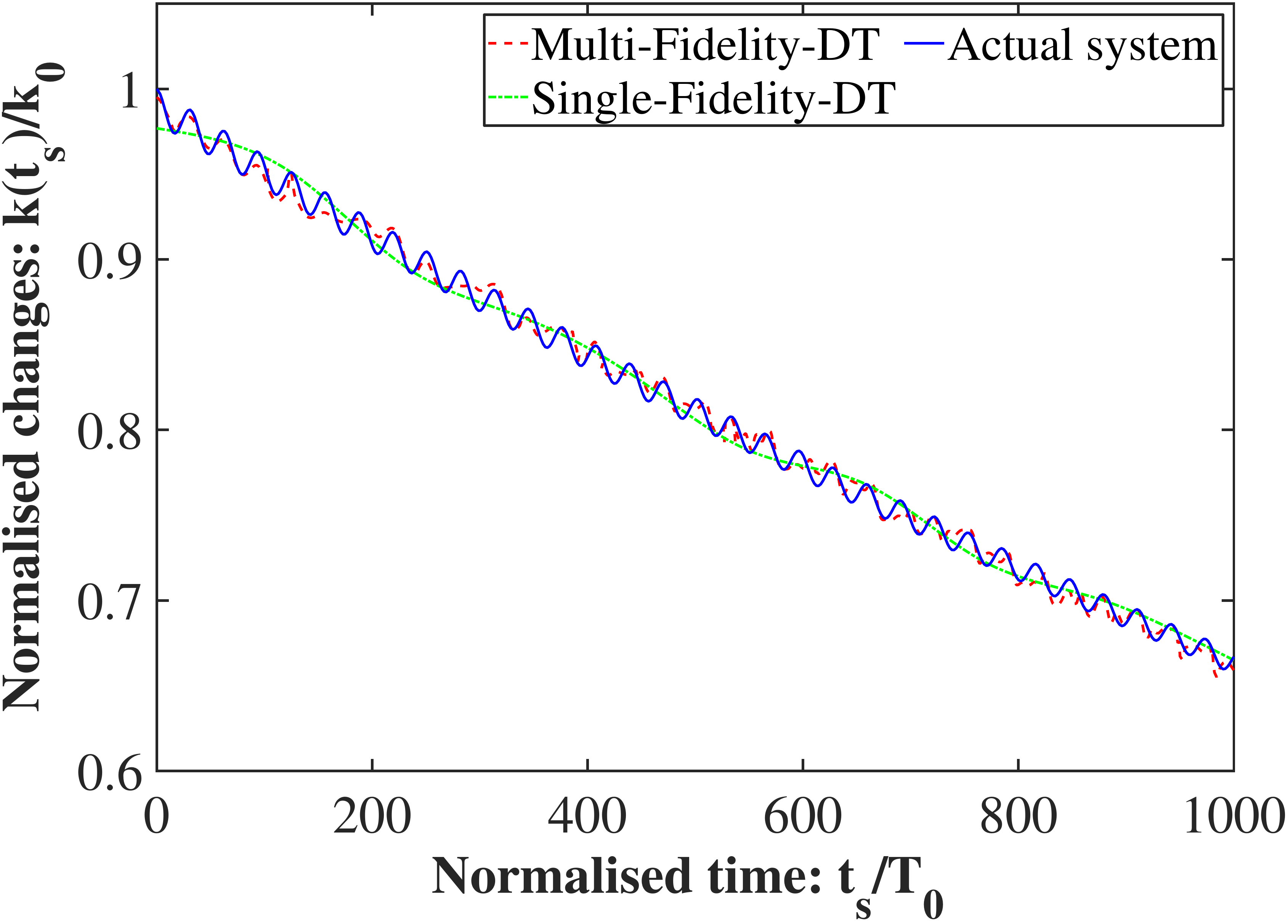}}
	\caption{
Prediction results of proposed MF-DT and Single-Fidelity-DT of stiffness evolution based on  frequency domain formulations:  Normalized response of the oscillator plotted against normalized time $\tau = t/T_0$. The results show the prediction of  MF-DT trained (a) with 3 HF and 501 data points, (b) with 4 HF and 501 data points, (c) with 5 HF and 501 LF data points, and (d) with 10 HF and 501 LF data points, where LF denotes low-fidelity (\autoref{m_fun}), and $HF_2$ denotes high-fidelity}\label{freqd_predk} 
\end{figure}

We further analyze the corresponding results of the stiffness degradation in \autoref{freqd_predk}(a-d). The results indicate the supremacy of the proposed framework. The multi-fidelity digital twin trained with 10 high-fidelity points and 501 low-fidelity points emulates the ground truth almost exactly. On the other hand, the single-fidelity digital twin 
yields poor performance in all cases. The results reinforce the consistency of the proposed multi-fidelity digital twin even with the frequency domain-based data processing.

\section{Conclusion}

The significance of digital twin technology is rapidly increasing in various engineering and industrial domains, including aerospace, infrastructure, and automotive. However, the practical implementation of digital twins faces challenges due to a lack of detailed application-specific information. To overcome this, data-driven models play a crucial role by integrating data and computational models, enabling real-time updates and predictions. Nevertheless, the efficiency of surrogate models, which bridge the physical systems and digital twin models, is often hindered by limited and inaccurate sensor data. An ideal surrogate mode for a digital twin should be able to incorporate data of multiple fidelity. To this end, we propose a novel framework that involves developing a robust multi-fidelity surrogate model, which can then be used for effective tracking of digital twin systems. Our proposed multi-fidelity modelling technique allows for the capture of nonlinear relationships between different fidelity data sets while eliminating erroneous information obtained from low-fidelity models. We introduce a framework called deep-Hybrid-Polynomial Correlated Function Expansion (deep-H-PCFE), which consists of nested individual H-PCFE models corresponding to different fidelity levels. Additionally, we refer to this framework as Multi-Fidelity H-PCFE (MF-H-PCFE) and Multi-Fidelity Digital Twin (MF-DT) throughout the paper. The basic building block of our framework is the H-PCFE, which combines Gaussian Process (GP) and Polynomial Correlated Function Expansion (PCFE) to efficiently approximate the actual function. Summarising, the key novel features of the proposed framework include:
\begin{enumerate}
    \item{The MF-H-PCFE proposed in this paper allows integrating data of multiple fidelity. Such as model is particularly useful for developing digital twin technology; this is because a realistic digital twin will have to incorporate multi-source and multi-fidelity data to update itself.}
    \item {The deep-HPCFE algorithm proposed in this paper is a Bayesian surrogate model and hence, provides an estimate of the predictive uncertainty. This, again, is a key feature toward enabling an uncertainty-aware digital twin framework. This, in the long run, will be helpful in taking an informed decision based on the digital twin predictions.}
    \item {The multi-fidelity surrogate model proposed in this paper is highly data efficient and requires only a few high-fidelity training samples. This can potentially be beneficial in developing digital twins from sparse and limited data.}
    \item {Last but not least, we have presented a framework that enables the development of digital twin technology from both time and frequency domain data. This is another major departure from our previous work on digital twin \cite{chakraborty2021role, chakraborty2021machine}}
\end{enumerate}
    
The proposed framework's applicability is demonstrated through numerical examples, starting with the validation of its performance in uncertainty quantification and, subsequently, its application in the development of digital twins from multi-fidelity data. The result obtained reinforces our claim and illustrates the advantages highlighted above. Overall, the proposed approach is a positive step towards the development of realistic digital twins of dynamical systems.

\section*{Acknowledgements}
NN acknowledges the support received from Ministry of Education in the form of Prime Minister's Research Fellowship, and A S Desai acknowledges the support received from the Ministry of Education through an M Tech. scholarship.  SC acknowledges the financial support of the Science and Engineering Research Board via grant no. SRG/2021/000467, Ministry of Port and Shipping via letter no. ST-14011/74/MT (356529) and seed grant received from IIT Delhi.

% \bibliographystyle{ieeetr}
% \bibliography{references}  %%% Uncomment this line and comment out the ``thebibliography'' section below to use the external .bib file (using bibtex).

\end{document}